\documentclass[10pt,twocolumn,letterpaper]{article}
\usepackage[dvipsnames]{xcolor}
\usepackage{subcaption}
\usepackage{float}
\usepackage{dblfloatfix}

\usepackage{wacv}
\usepackage{times}
\usepackage{epsfig}
\usepackage{graphicx}
\usepackage{abstract}
\usepackage{amsmath}
\usepackage{amssymb}
\usepackage{booktabs}

\usepackage{amsmath,amsfonts,bm}

\newcommand{\myparagraph}[1]{\vspace{0pt}{\noindent\bf #1}}

\def\1{\bm{1}}

\def\sp{space}

\def\vc{{\bm{c}}}

\def\ve{{\bm{e}}}

\def\vt{{\bm{t}}}

\def\vw{{\bm{w}}}
\def\vx{{\bm{x}}}

\def\vz{{\bm{z}}}

\def\mI{{\bm{I}}}

\DeclareMathAlphabet{\mathsfit}{\encodingdefault}{\sfdefault}{m}{sl}
\SetMathAlphabet{\mathsfit}{bold}{\encodingdefault}{\sfdefault}{bx}{n}

\def\gC{{\mathcal{C}}}
\def\gD{{\mathcal{D}}}

\def\gG{{\mathcal{G}}}

\def\gN{{\mathcal{N}}}

\usepackage{comment}
\usepackage{color}
\usepackage{symbols}
\usepackage{multirow}
\usepackage{makecell}
\usepackage{ctable}
\usepackage{cite}
\usepackage{enumitem}
\usepackage{authblk}

\usepackage{bbding}

\definecolor{darkred}{rgb}{0.7,0.1,0.1}
\definecolor{darkgreen}{rgb}{0.1,0.7,0.1}
\definecolor{cyan}{rgb}{0.7,0.0,0.7}
\definecolor{dblue}{rgb}{0.2,0.2,0.8}
\definecolor{maroon}{rgb}{0.76,.13,.28}
\definecolor{burntorange}{rgb}{0.81,.33,0}
\definecolor{tealblue}{rgb}{0.212,0.459, 0.533}

\definecolor{pp}{rgb}{0.43921569, 0.18823529, 0.62745098}
\definecolor{rr}{rgb}{0.5254902 , 0.00784314, 0.12941176}
\definecolor{bb}{rgb}{0.09019608, 0.23529412, 0.37647059}
\definecolor{yy}{rgb}{0.49803922, 0.3372549 , 0.0}
\definecolor{gg}{rgb}{0.02352941, 0.3372549 , 0.17647059}

\usepackage{xspace}
\newcommand{\modelname}{\textit{Fast text2StyleGAN}\xspace}
\newcolumntype{C}[1]{>{\centering\arraybackslash}p{#1}}

\usepackage{authblk}
\makeatletter
\renewcommand\AB@affilsepx{ \protect\Affilfont}
\makeatother

\def\wacvPaperID{668} %

\wacvalgorithmstrack   %

\wacvfinalcopy %

\usepackage[colorlinks=true,citecolor=NavyBlue,linkcolor=OrangeRed,urlcolor=Rhodamine]{hyperref}

\begin{document}

\title{Text-Free Learning of a Natural Language Interface\\ for Pretrained Face Generators}
\author[1]{Xiaodan Du\textsuperscript{1}, Raymond A. Yeh\textsuperscript{2}\thanks{Work done at TTI-Chicago}, Nicholas Kolkin\textsuperscript{3}, Eli Shechtman\textsuperscript{3}, Greg Shakhnarovich}
\affil[1]{TTI-Chicago{\tt\small \{xdu,gregory\}@ttic.edu}\;\;\;\;\;\;\;\;
\textsuperscript{2}Purdue University{\tt\small \{rayyeh\}@purdue.edu}
\newline \textsuperscript{3}Adobe Research{\tt\small \{kolkin,elishe\}@adobe.com}
}

\twocolumn[{%
	\renewcommand\twocolumn[1][]{#1}%
	\maketitle
	\begin{minipage}{\textwidth}
\vspace{-0.6cm}
\centering
\setlength{\tabcolsep}{0pt}
\begin{tabular}{ccc@{\hspace{1.5em}}ccc}

\multicolumn{3}{c}{\small A photo of a young person.} & \multicolumn{3}{c}{\small A photo of a young person with long curly hair.}\\

\includegraphics[width = .16\textwidth]{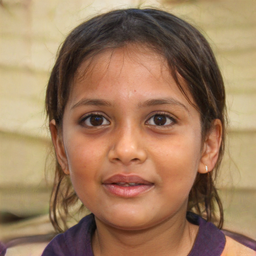}&
\includegraphics[width = .16\textwidth]{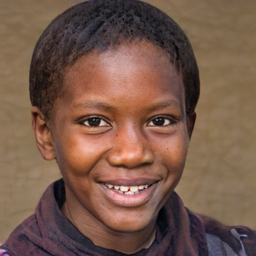}&
\includegraphics[width = .16\textwidth]{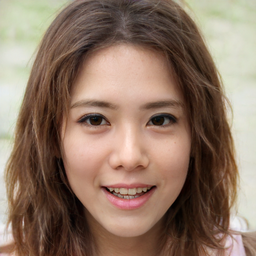}&

\includegraphics[width = .16\textwidth]{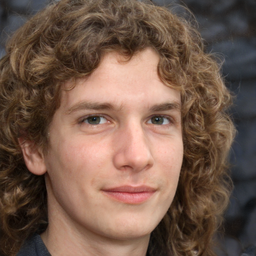}&
\includegraphics[width = .16\textwidth]{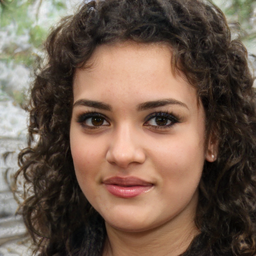}&
\includegraphics[width = .16\textwidth]{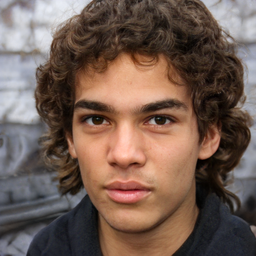}\\

\multicolumn{3}{C{0.45\textwidth}}{
\begin{tabular}{l}
\small A photo of a laughing young person with long curly hair.
\end{tabular}
} & \multicolumn{3}{C{0.45\textwidth}}{
\small 
\begin{tabular}{l}
A photo of a laughing young person with long curly hair\\ wearing glasses.
\end{tabular}
}\\

\includegraphics[width = .16\textwidth]{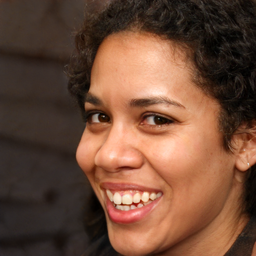}&
\includegraphics[width = .16\textwidth]{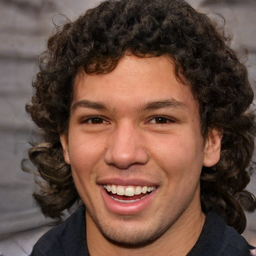}&
\includegraphics[width = .16\textwidth]{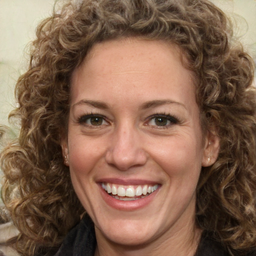}&

\includegraphics[width = .16\textwidth]{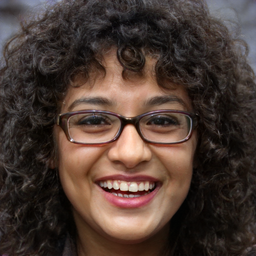}&
\includegraphics[width = .16\textwidth]{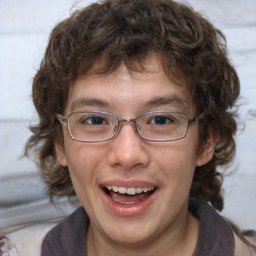}&
\includegraphics[width = .16\textwidth]{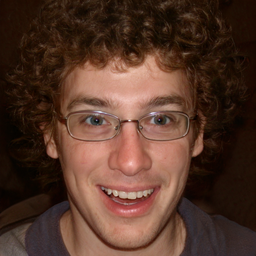}
\\
\end{tabular}
\captionof{figure}{Examples of text-driven face image synthesis by our proposed method \modelname. The text prompts are increasingly more detailed. Each image takes about 0.09s to produce. 
See {\footnotesize \url{https://github.com/duxiaodan/Fast_text2StyleGAN.git}}
\vspace{0.3cm}}
\label{fig:teaser_new2}
\end{minipage}

}]
\saythanks

\thispagestyle{empty}

\begin{abstract}
\vspace{-0.2cm}
We propose~\modelname, a natural language interface that adapts pre-trained GANs for text-guided human face synthesis. Leveraging the recent advances in Contrastive Language-Image Pre-training (CLIP), no text data is required during training.~\modelname is formulated as a conditional variational autoencoder (CVAE) that provides extra control and diversity to the generated images at test time. Our model does not require re-training or fine-tuning of the GANs or CLIP when encountering new text prompts. In contrast to prior work, we do not rely on optimization at test time, making our method orders of magnitude faster than prior work. Empirically, on FFHQ dataset, our method offers faster and more accurate generation of images from natural language descriptions with varying levels of detail compared to prior work.

\vspace{-0.3cm}
\end{abstract}

\begin{figure*}
    \centering
    \includegraphics[width=\linewidth]{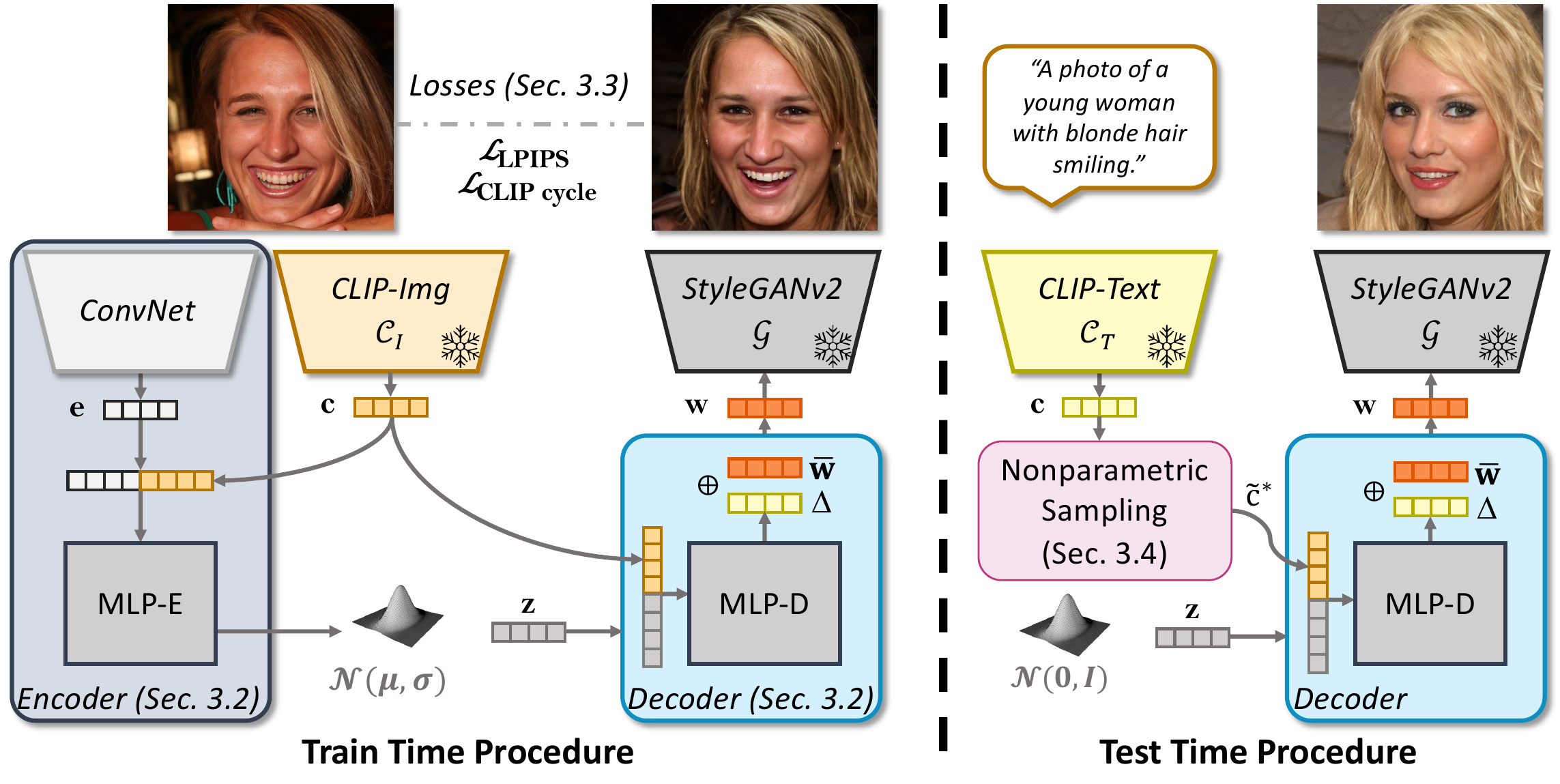} 
    \caption{Illustration of our~\modelname model. At train time, the model parameters of CLIP's image encoder and StyleGAN components are frozen. At test time, the CLIP image embedding is swapped out with a CLIP text embedding. We perform non-parametric sampling to mitigate the mismatch caused by this swap. Only a single forward pass is needed for generation. \raisebox{-.2em}{\SnowflakeChevron}~indicates pre-trained and ``frozen" components.}
    \label{fig:architecture}
    \vspace{-0.5cm}
\end{figure*}
\section{Introduction}\label{sec:intro}
Modern generative models like StyleGANv2 can map a random latent code to a photorealistic image, especially in the human face domain. In many application scenarios, to be useful in practice, the process of generation must allow for explicit control of the content. Previous work explored modes of control based on analysis of a pre-trained model and identification of domain-specific axes of variation in the GAN latent space (\eg, age or expression for faces). Then, by changing the values of the latent variables along these axes, one can modify the generated image. These techniques are helpful for editing an image, but for complete de-novo synthesis they are inefficient.

To address this issue, we pursue a more natural control mechanism: via explicit natural language description of the desired properties of the generated content. In other words, we aim to generate \textit{human face domain-specific} photorealistic images purely from natural language descriptions; See~\figref{fig:teaser_new2}.
This task is often formulated as learning a conditional generative model, for which a dataset of paired text and images is required during training. However, many image datasets do not come with natural descriptions.
Naively, one could annotate the images manually but it is labor intensive and time consuming, and the annotation protocol is ambiguous and arbitrary. Generating a representative dataset for learning a flexible mapping of descriptions to images would require many complementary descriptions per image, or a very large dataset, or both~\cite{ramesh2022hierarchical}. %
In this work, we present an alternative solution that does not require \emph{any} text/image pairs in the target domain.

Our solution involves learning an interface between two components: (a) A pre-trained image generator, which can be trained from the domain specific images; we use StyleGANv2~\cite{karras2020analyzing}. (b) Multi-modal (text and image) embedding models trained on very large, non-domain-specific association corpora, specifically CLIP~\cite{radford2021learning}.
At a high-level, we train an ``adapter" from CLIP's image/text embedding space to the latent space of an image generator. This training is guided by (approximate) image reconstruction as the main objective. Once this adapter is trained, we can feed it the CLIP embedding of a text (the description of the desired output) instead of the CLIP embedding of an image. To the extent that CLIP image and text embeddings are interchangeable, this allows us to generate an image from text without ever seeing an image/text pair during training of our interface. See overview in~\figref{fig:architecture}.

A key technical difficulty, however, is that CLIP embeddings of images do not necessarily capture all the perceptually  important aspects of the image. For instance, we find that while the coarse ``demographics" of the face are well captured by CLIP, the pose (orientation) and background are not. In addition, a description may omit certain visual attributes (\ie, it may refer to the desired expression and age, but not to the hair style). We would like to be able to achieve diversity in these aspects that are under-specified by the CLIP embedding of the description. To this end we propose a novel two-channel encoder for images. One channel is the (pre-trained, fixed) CLIP embedding. The other channel is a conditional variational autoencoder (CVAE), conditioned on the CLIP embedding. The CVAE channel, intuitively, learns to complement the information captured by CLIP, to allow better image reconstruction at training time, and to provide a source of diversity (and possibly additional control) in the generated images at inference time.  

A remaining issue is the discrepancy of CLIP text vs. image embedding distribution, and the lack of diversity within attributes that \emph{are} specified by the description. Thus, as an additional contribution, we propose a non-parametric prior for mapping text embedding to a distribution of image embeddings, without training. %

To date most successful GAN models are trained for one domain at a time, and they have achieved particular success on the domain of human faces, where they can produce diverse images of synthetic faces at very high resolutions. This, and the importance of faces in applications (\eg, synthetic stock photos) motivates us to focus on faces as our main experimental domain. Empirically, we demonstrate that our approach generates high quality images and more faithfulness to text prompt than prior work. Importantly, our method can generate images that follows complex language description with different level of details. Additionally, our approach is very fast and requires only a single forward pass to generate an image, hence we name it~\modelname.

{\noindent \bf Our contributions are as follows:}
\begin{itemize}[topsep=0pt,itemsep=-1ex,partopsep=1ex,parsep=1ex,leftmargin=*]
    \item We propose \modelname, a natural language interface between a joint image/text embedding model and a GAN. We introduce a novel two-channel architecture, proposing a CLIP-conditional variational autoencoder for training this interface.
    \item \modelname generates high resolution images from text dramatically (orders of magnitude) faster than prior work. 
    \item Our method supports arbitrary text inputs at inference time; \textit{no re-training or fine-tuning on paired image and text prompt is necessary.} %
    \item We propose a non-parametric %
    text-to-image embedding prior which significantly improves diversity and accuracy.
\end{itemize}

\section{Related Work}\label{sec:related}

{\bf\noindent CLIP.} By leveraging a massive dataset of 400 million text-image pairs, CLIP~\cite{radford2021learning} uses contrastive loss to learn a multi-modal embedding of images and text into a shared ``semantic space". Associated image-text pairs are mapped to vectors with higher cosine similarity than random pairs. These embeddings have proven to be broadly applicable, becoming a vital component in a variety of image generation and manipulation frameworks~\cite{Patashnik_2021_ICCV, ramesh2022hierarchical, gal2021stylegan}.

\myparagraph{Unconditional image generation.} While unconditional image generation remains an open problem, four leading paradigms have emerged in recent years: autoregressive models~\cite{van2016pixel,esser2021imagebart,de2019hierarchical}, VAE models\cite{van2017neural,razavi2019generating}, diffusion models~\cite{song2019generative, ho2020denoising}, and GANs~\cite{goodfellow14, miyato2018spectral}. Of these GANs are perhaps the most popular, owing to samples' high visual quality, fast inference speed, and the ease of performing impressive image manipulations using their fairly compact latent space. We adopt StyleGANv2~\cite{karras2020analyzing}, which at inference time maps a Gaussian random input $\mathbf{z}$ to a transformed latent vector $\mathbf{w}$, then decodes $\mathbf{w}$ into an image.

\myparagraph{Image manipulation in latent space.} A great deal of recent work has focused on ``inverting'' a generator, typically a GAN,~\ie, to find a point in the latent space that, when fed to the generator, generates a given image~\cite{richardson2021encoding, tov2021designing,abdal2019image2stylegan,alaluf2021restyle}, then performing image manipulations by altering the latent code. A major line of work in this area focuses on discovering dimensions of the latent space which control certain predetermined characteristics of the image (most commonly, attributes of face images, such as age, gender, expression, hairstyle~\etc~\cite{wu2021stylespace,shen2020interfacegan,zhu2021barbershop}). An input image is ``inverted", \ie, mapped to the latent space, the latent representation is moved along relevant attribute-specific dimensions, and fed back to the generator which produces an image with the desired attributes modified. This approach has simplified complex image manipulations which only a few years ago would have required time consuming effort of experts. However, these methods are limited by requiring controls in latent space to be found before inference, and lack the flexibility and intuitive control of natural language.

\myparagraph{Text-driven image generation.}
Image manipulation is just one aspect of control which increases the utility of generative models, and conditional de-novo generation is another way generative models can become a useful tool. This second regime, specifically text-driven conditional generation, is our focus. GANs~\cite{xu2018attngan,li2019object,tao2022df,zhang2021cross} have been widely used for text-driven image synthesis tasks. However, unlike most work in this area, \modelname does not rely on additional labels such as captions, bounding boxes, or segmentation maps; instead using the CLIP embedding of images in the training set as a proxy for paired captions. Two recent works operate in a similar regime, TediGAN~\cite{xia2021tedigan} and StyleCLIP~\cite{Patashnik_2021_ICCV}, making them a good point of comparison with~\modelname. We conduct qualitative and quantitative comparisons with these works. %

Recently several works have made great strides in generating high quality images conditioned on natural language. DALL-E~\cite{ramesh2021zero, ramesh2022hierarchical}, CogView~\cite{ding2021cogview, ding2022cogview2}, ImaGen~\cite{imagen},~\etc~\cite{StyleT2I,anyface}, have drawn a lot of attention because of their ability to synthesize diverse and high-quality images from long and complex prompts. However, they usually {\it require large amount of  paired training data} and computing resources.
In this work, we do not compare to these models because: (a) they require text data during training; and/or (b) they are trained on much larger domain than human face so it is hard to compare quantitatively; especially when most of them are not open-sourced.

\section{Approach}
The overview of our approach is shown in~\figref{fig:architecture}. We rely on two pre-trained (and frozen) components: the StyleGANv2 image generator $\mathcal{G}$, and the CLIP image/text embedding networks, respectively $\mathcal{C}_{I}$ and $\mathcal{C}_{T}$. The end goal is to map $\mathcal{C}_T(\text{prompt})$ to the latent space of $\mathcal{G}$. We do so without any availability of detailed prompt/image pairs. Leveraging the multi-modal nature of CLIP, we formulate the problem as a conditional variational autoencoder CVAE~\cite{sohn2015learning} in~\secref{sec:cvae}. Architecture and training details are discussed in~\secref{sec:arch} and~\secref{sec:losses}.

{\bf At train time}, an image is encoded into two parts: (a) a CLIP image embedding for conditioning both the encoder and decoder in CVAE; (b) a latent vector that encodes the variation of the image to follow a Gaussian distribution. Given these two vectors, the decoder's aims to reconstruct the image. {\bf At test time}, our goal is to perform text conditional generation. Given a text input, we extract a text embedding using CLIP. 
Due to the distribution gap between CLIP's text/image embedding, we construct a non-parametric generative model of CLIP image embedding given CLIP text embedding; see~\secref{sec:knn}. Using this model, we sample a corresponding CLIP image embedding to condition the decoder. Finally, we sample the latent vector from a Gaussian distribution following a standard CVAE. Importantly, our approach requires a single forward pass through the decoder at test time leading to a short generation time, hence we named our model~\modelname. 

\subsection{Interface learning with CVAE}\label{sec:cvae}
Let $\vx$ denote an image. Our goal can be viewed as modeling a conditional distribution $p(\vx|\vc)$, where $\vc$ is a conditioning \emph{context}. At training time, $\vc$ is provided by the (deterministic) CLIP-img embedding $\vc(\vx)=\mathcal{C}_I(\vx)$.  CVAE incorporates both inference (mapping from image to the latent) and generation (mapping from latent to image). Given an input $\vx$ and a \emph{context} $\vc$, CVAE maps them to two components: $f_{\tt enc}^\mu$ and $f_{\tt enc}^{\sigma^2}$, that models the mean and (per-dimension) variance of the Gaussian distribution for a latent vector $\vz$. To enable sampling, the distribution of the latent is required to match the prior:
\bea
q_\phi(\vz | \vx, \vc) =&\gN(f_{\tt enc}^\mu(\vx, \vc), f_{\tt enc}^{\sigma^2}(\vx, \vc))\\
\label{eq:generation}
p_\theta(\vx | \vz, \vc) =& \gN(f_{\tt dec}^\mu(\vz, \vc), \sigma\mI)\\
p(\vz | \vc) =& \gN(\mathbf{0},\mathbf{I}).
\eea
Typically, the prior is chosen to follow a standard multivariate Gaussian.
A standard CVAE is trained on an image set $\mathcal{D}=\{\vx\}$ by maximizing the evidence lower
bound (ELBO):
\bea\nonumber
\hspace{-3em}\sum_{\hspace{3em}(\vx,\vc(\vx)) \in \gD} \hspace{-3em}\sE_{q_\phi{(\vz|\vx,\vc)}} \left[
\log(p_\theta(\vx | \vz, \vc)) - D_{\text{KL}}\left(q_\phi(\vz|\vx,\vc)\| p(\vz|\vc)\right)
) \right].
\eea
In practice, additional loss functions were introduced to improve image generation quality; we defer discussion of these to~\secref{sec:losses}. We will next describe the model's architecture.

\subsection{CVAE architecture}\label{sec:arch}
The overall architecture of~\modelname can be viewed as a CVAE shown in~\figref{fig:architecture}. The encoder parameterizes the distribution $q_\phi$, where as the decoder parameterizes the distribution $p_\theta$. At test-time time, given the text description $\vt$, a random vector drawn from standard Gaussian distribution $\mathcal{N}(0,I)$ and the CLIP embedding of the text prompt $\vc=\mathcal{C}_{T}(\vt)$ are used as inputs.
During inference time, an image, instead of a random noise, can be used as extra guidance in addition to text prompt. %

\myparagraph{Encoder $q_\phi(\vz|\vx, \vc)$ architecture.}
Our CVAE's encoder consists of two branches in parallel. Given an image as input the first branch extracts the CLIP embedding via CLIP's pre-trained encoder, $\vc = \mathcal{C}_I(\vx)$. Another branch in parallel is a standard convolutional neural network containing five 3$\times$3 \texttt{Conv2d} layers and a fully-connected layer. Let its output be $\ve = \textit{ConvNet}(\vx)$. 

The two output vectors are concatenated, and mapped by a four-layer network $\text{MLP-E}$ to output the the mean and variance of the latent space: 
\bea\label{eq:mlpe}
[f_{\tt enc}^\mu, f_{\tt enc}^{\sigma^2}] = \text{MLP-E}([\ve, \vc]).
\eea
To sample latent vector $\vz$ from the posterior distribution $q_\phi(\vz|\vx, \vc)$ and back-propagate through the samples for training, we use the reparameterization trick~\cite{kingma2013auto}.

\myparagraph{Decoder $p_\theta(\vx|\vz,\vc)$ architecture.}
The CLIP embedding of the input $\vc$ is concatenated with $\vz$ and sent into a decoder network $\text{MLP-D}$, with architecture identical to that of $\text{MLP-E}$. $\text{MLP-D}$ outputs vector $\Delta$, interpreted as the offset from the average latent code $\bar\vw$ of StyleGAN's latent. We follow  definition of $\bar\vw$ by Richardson~\etal~\cite{richardson2021encoding}. Finally, $\Delta$ is added to $\bar{\vw}$ and passed to StyleGANv2, $\gG$, to produce the output image
\bea
\hat{\vx} = \gG(\bar{\vw}+\Delta), ~\text{where}~ \Delta = \text{MLP-D}([\vz,\vc]).
\eea

\subsection{Loss functions}\label{sec:losses}
Inspired by Richardson~\etal~\cite{richardson2021encoding} and Tov~\etal~\cite{tov2021designing}, we replaced the pixel-wise $\ell_2$-loss resulting from the Gaussian assumption (\equref{eq:generation}) with the Learned Perceptual Image Patch Similarity \textbf{(LPIPS) loss}~\cite{zhang2018perceptual} to ensure high perceptual similarity between the deep representations of input and output images.

Next, to encourage our model to learn input and output images that are similar in CLIP space (to eventually drive text generation) we introduce a 
\textbf{CLIP cycle loss}:
\bea
    \mathcal{L}_{\text{CLIP cycle}}(\vx) = 1- \text{Sim}_{\text{cos}}\left(
    \mathcal{C}_I(\vx), \mathcal{C}_I\left(\hat{\vx}\right)
    \right),
\eea
where $\text{Sim}_{\text{cos}}$ represents the cosine similarity between the two embeddings and $\gC_I$ is the image encoder of CLIP. Recall (Sec.~\ref{sec:arch}) that during training $\Delta$ is the output of MLP-D, computed from both the CLIP embedding of $\vx$ and from $\vz$, sampled from the output of the variational encoder.

Finally, following Richardson~\etal~\cite{richardson2021encoding}, we included a \textbf{$\vw$ normalization loss} to prevent the predicted StyleGAN latent vector $\vw$ from straying too much from the distribution:
\bea
    \mathcal{L}_{\vw\text{-norm}}(\vx) = \| \Delta\|^2.
\eea
To train the model, we freeze model parameters of CLIP and StyleGANv2 and only perform gradient updates on MLP-E and MLP-D to minimize a weighted combination of the aforementioned losses.

\subsection{Non-parametric sampling for text-to-image embedding generation}\label{sec:knn}
At test-time, we are given a text $\vt$ of the desired image. Naively, one can directly use this text's CLIP embedding,  $\mathcal{C}_T(\vt)$, in-place of an image embedding for conditioning the decoder. However, this CLIP embedding only provides a deterministic descriptor for the content, with the CVAE Gaussian latent sample $\vz$ being only source of randomness. 

This poses two problems. First, while CLIP's training objective is to maximize cosine similarity between associated text/image pairs, there is a significant discrepancy between image and text embeddings, yielding a discrepancy between training and inference regime, potentially reducing generation accuracy. Second, even a relatively detailed description of an image necessarily leaves a lot unspecified; while sampling $\vz$ conditional on $\mathcal{C}_T(\vt)$ provides some of these complementary details, we find that the resulting diversity is too limited, as seen in~\figref{fig:gaussian_knn_compare_new}. 

A similar observation in concurrent text-to-image generation work, notably in DALL-E2~\cite{ramesh2022hierarchical}, where the solution is to train %
a 1B parameter model that maps CLIP text embedding $\mathcal{C}_T(\vt)$ to a distribution of CLIP image embeddings $\mathcal{C}_I(\vx)$ for images that match $\vt$. This requires a dataset of image/text pairs, which in DALL-E2 count in the hundreds of millions. Our approach is aimed at domains where we do not have \emph{any} paired image/text data. Thus, we propose a very simple, and as we show very effective, non-parametric conditional generative model that does not require separate training and uses domain images only.

At test time, we retrieve $K$ training images whose CLIP image embeddings, $\vc^\ast_1,\ldots,\vc^\ast_K$ are closest in cosine distance to the CLIP text embedding of the prompt, $\mathcal{C}_T(\vt)$. Then we sample a random subset of $M<K$ of these nearest neighbors -- without loss of generality, let these be $\vc^\ast_1,\ldots,\vc^\ast_M$. We compute a random convex combination $\tilde{\vc}^\ast\,=\,\sum_{j=1}^M\alpha_j\vc^\ast_j$, with $\alpha_j$s sampled from a Dirichlet distribution. 
The resulting vector $\tilde{\vc}^\ast$, instead of $\mathcal{C}_T(\vt)$, is 
passed to the decoder at test time.
We observe the non-parametric model to yield better visual quality, presumably because $\vc$ at test time is closer to the $\vc$ used at training time. The stochasticity in the process (random $M$ neighbors and random convex combination of their embeddings) also tends to provide more meaningful and significant diversity. Visual comparisons can be found in~\figref{fig:gaussian_knn_compare_new}. %

\section{Experiments}\label{sec:exp}
We conduct experiments using the Flickr-Faces-HQ (FFHQ) \cite{karras2019style} dataset. FFHQ contains 70,000 diverse 1024$\times$1024 high-resolution images of human faces, which StyleGANv2 is also trained on.

\begin{table}[t]
\setlength{\tabcolsep}{1pt}
\small
\resizebox{\linewidth}{!}{
\centering
\begin{tabular}{ccccccc}
\specialrule{.15em}{.05em}{.05em}
\multicolumn{1}{c}{}
 & \multicolumn{4}{c}{Correctness (retrieval accuracy)}
 & \multicolumn{1}{c}{Diversity} \\
\cmidrule(lr){2-5} %
\cmidrule(lr){6-6}
 Models %
 & Top 1$\uparrow$      & Top 5$\uparrow$     & Top 10$\uparrow$     & Top 20$\uparrow$  & ID Div.$\uparrow$  & Time$\downarrow$\\
\hline
\hline

StyleCLIP (m)
 & 0.148       & 0.341      & 0.463       & 0.619       &  0.930       & 55s
\\
TediGAN (m)         %
& 0.380       & 0.608      & 0.731       & 0.835     & 0.903      &  21s\\
StyleCLIP (g) 
& 0.767       & 0.941      & 0.978       & 0.992      & 0.319        & 55 s
\\
TediGAN (g)         %
& 0.346       & 0.589      & 0.721       & 0.840      & 0.818          &  21s\\

\hline
Ours (Pt) %
& 0.690       & 0.875      & 0.939       & 0.984     & 0.345     &  0.018s     \\
Ours (Img)      %
& 0.706       & 0.901      & 0.940       & 0.975     & 0.357     & 0.019s \\
Ours     %
&  0.795       &  0.974      &  0.993       &  1.000     & 0.495      & 0.094s \\
\specialrule{.15em}{.05em}{.05em}
\end{tabular}}
\caption{Quantitative comparison of our models and prior works. Our method produces images that correctly follows the text, with diversity, and is much faster than the baselines. %
}
\vspace{-0.5cm}
\label{tab: results}
\end{table} 
\myparagraph{Evaluation metrics.} For empirical validation, we consider three aspects, image quality, image diversity, and correctness. Following the protocol of prior work~\cite{liu2021more}, we use their method to generate 400 captions. All models randomly generate 25 images for each caption. The 10,000 generated images for each model are used for comparison.

 {\it For quality}, we have included numerous non-curated examples (also in the Appendix) generate from our approach and baseline. We refrain from using FID~\cite{heusel2017gans} for evaluation as the metric is designed for unconditional generation. We provide a detailed discussion and evidence on why FID is not suitable for %
 conditional generation in the Appendix. 
 
 {\it For correctness}, we report retrieval accuracy which measures the consistency with input text. For an output image $\hat{\vx}$ with text guidance $\vt$, 99 images in FFHQ dataset are randomly chosen as negative samples. CLIP is used to measure the cosine similarity between $t$ and the 100 images. Retrieval accuracy is the average over all text prompts and generated images. This metric is introduced in a prior work~\cite{liu2021more}.
  
 {\it For diversity}, we report identity diversity ({\it ID Div.}) %
 Recall for each of the 400 captions we generate 25 images resulting %
 in ${25 \choose 2} = 300$ image pairs. For each image pairs, we use the pre-trained ArcFace~\cite{arcface} model to obtain feature vectors for both images. We then compute the cosine similarity between these two vectors. We denote one minus this similarity as the identity diversity. Finally, we average the identity diversity over all image pairs and all captions. As ArcFace is trained to perform face recognition, features extracted from ArcFace focus on the facial attributes.~\Ie, a high {\it ID Div.} indicates that the two face images are not of the same identity, hence, more diverse.

\begin{figure*}[ht]
\centering
\setlength{\tabcolsep}{0pt}
\begin{tabular}{
>{\begin{sideways}}c<{\end{sideways}}
ccc@{\hspace{0.3em}}ccc@{\hspace{0.3em}}ccc}

& \multicolumn{3}{c}{ \small 
    \begin{tabular}{c}
    (a) He is old.
  \end{tabular}
  } &\multicolumn{3}{c}{ \small 
  \begin{tabular}{c}
   (b) A blond Asian girl.
  \end{tabular}
  } &\multicolumn{3}{C{.33\textwidth}}{ \small 
  \begin{tabular}{c}
  (c) An old Black woman wearing glasses,\\ with a thoughtful expression on her face.
  \end{tabular}
  } \\

{\small StyleCLIP (m)} & \includegraphics[width = .11\textwidth]{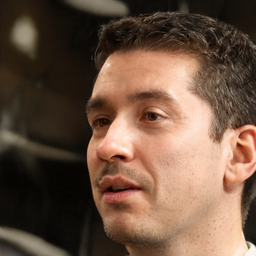}&
 \includegraphics[width = .11\textwidth]{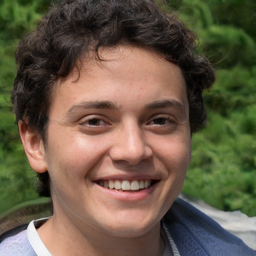}&
 \includegraphics[width = .11\textwidth]{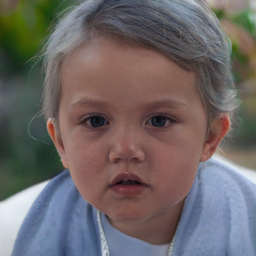}&
\includegraphics[width = .11\textwidth]{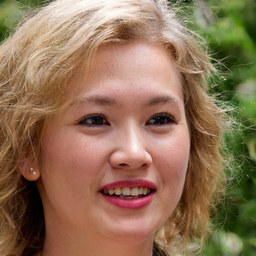}&
 \includegraphics[width = .11\textwidth]{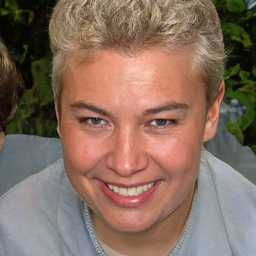}&
 \includegraphics[width = .11\textwidth]{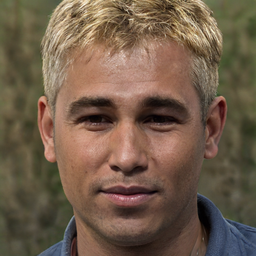}
 &\includegraphics[width = .11\textwidth]{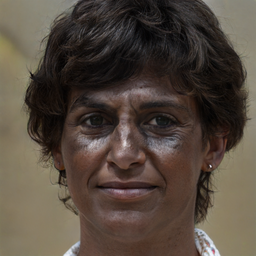}&
 \includegraphics[width = .11\textwidth]{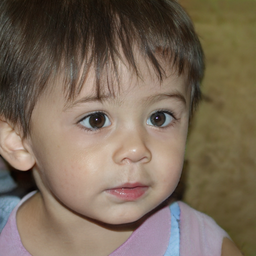}&
 \includegraphics[width = .11\textwidth]{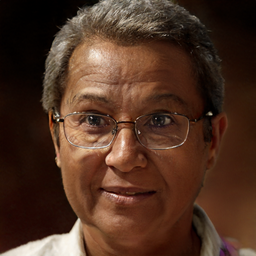}\\
 
{\small TediGAN (m)} & \includegraphics[width = .11\textwidth]{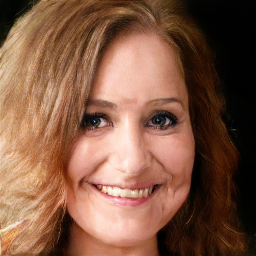}&
 \includegraphics[width = .11\textwidth]{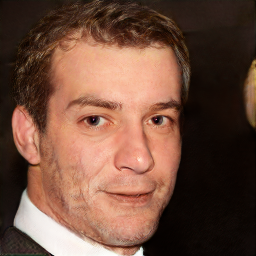}&
 \includegraphics[width = .11\textwidth]{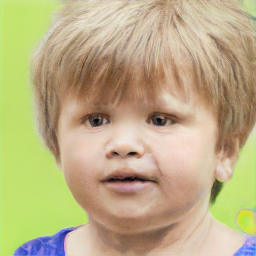}&
\includegraphics[width = .11\textwidth]{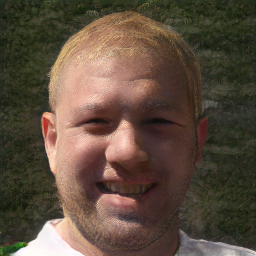}&
 \includegraphics[width = .11\textwidth]{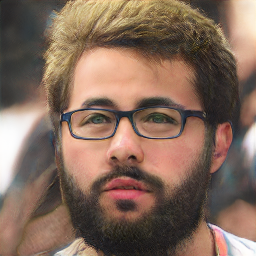}&
 \includegraphics[width = .11\textwidth]{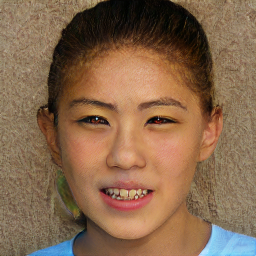}
 &\includegraphics[width = .11\textwidth]{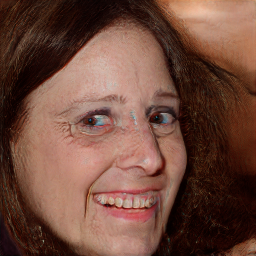}&
 \includegraphics[width = .11\textwidth]{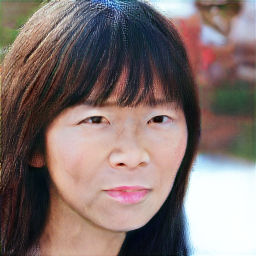}&
 \includegraphics[width = .11\textwidth]{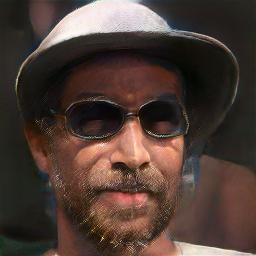}\\
{\small StyleCLIP (g) }& \includegraphics[width = .11\textwidth]{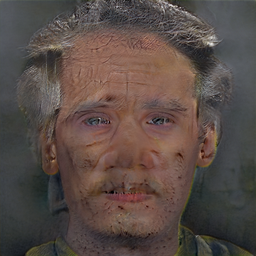}&
 \includegraphics[width = .11\textwidth]{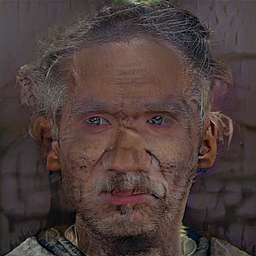}&
 \includegraphics[width = .11\textwidth]{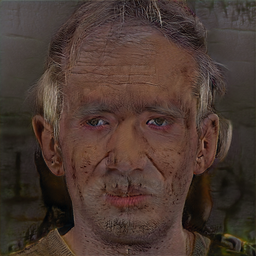}&
\includegraphics[width = .11\textwidth]{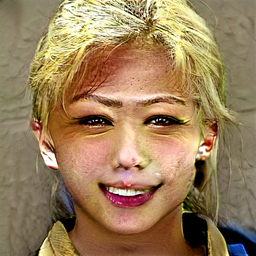}&
 \includegraphics[width = .11\textwidth]{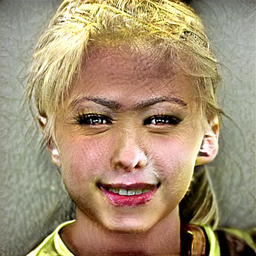}&
 \includegraphics[width = .11\textwidth]{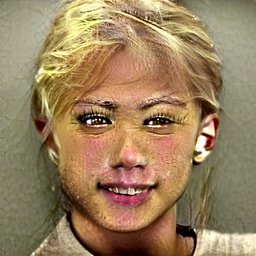}
 &\includegraphics[width = .11\textwidth]{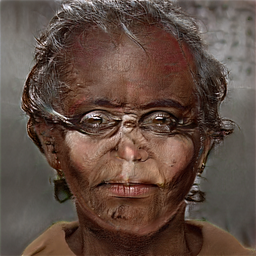}&
 \includegraphics[width = .11\textwidth]{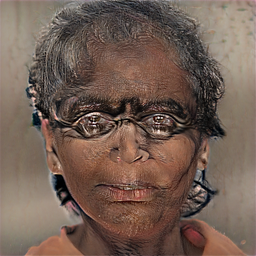}&
 \includegraphics[width = .11\textwidth]{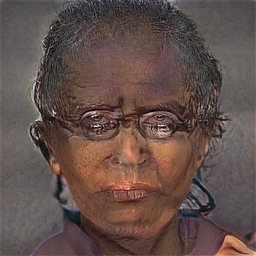}\\
{\small TediGAN (g)} & \includegraphics[width = .11\textwidth]{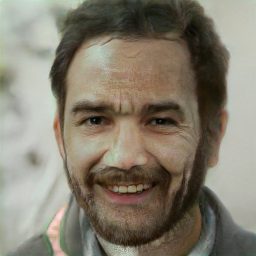}&
 \includegraphics[width = .11\textwidth]{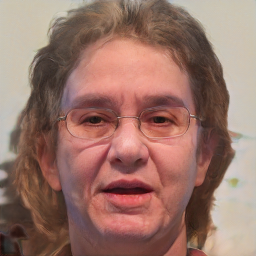}&
 \includegraphics[width = .11\textwidth]{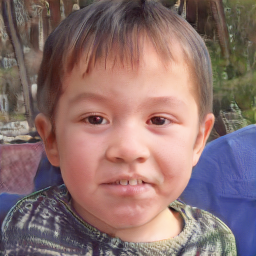}&
\includegraphics[width = .11\textwidth]{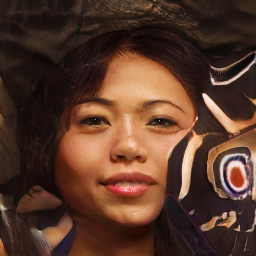}&
 \includegraphics[width = .11\textwidth]{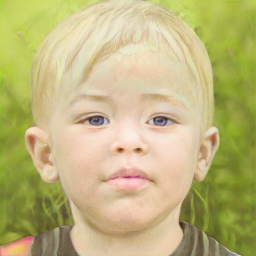}&
 \includegraphics[width = .11\textwidth]{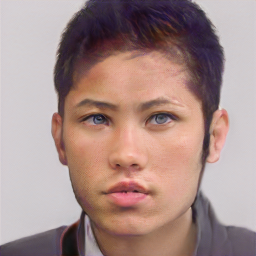}
 &\includegraphics[width = .11\textwidth]{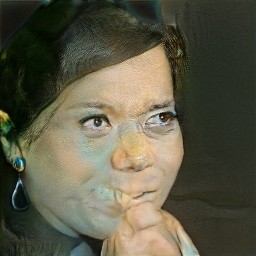}&
 \includegraphics[width = .11\textwidth]{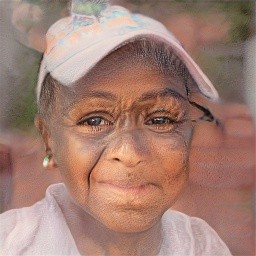}&
 \includegraphics[width = .11\textwidth]{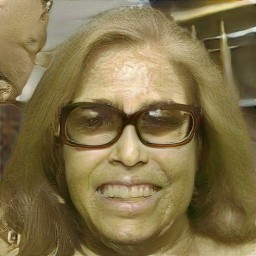}\\

\hspace{0.55cm}\textbf{Ours} & \includegraphics[width = .11\textwidth]{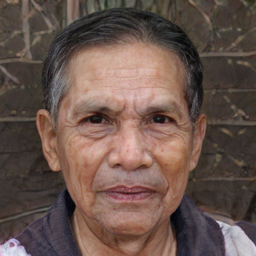}&
 \includegraphics[width = .11\textwidth]{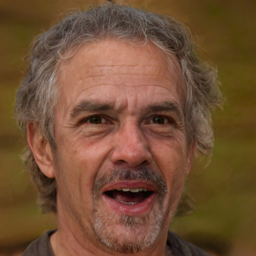}&
 \includegraphics[width = .11\textwidth]{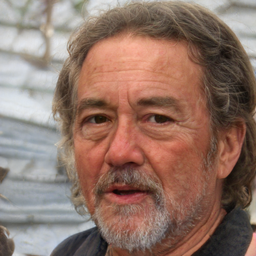}&
\includegraphics[width = .11\textwidth]{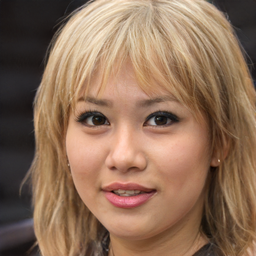}&
 \includegraphics[width = .11\textwidth]{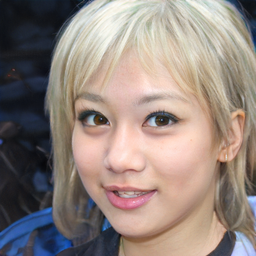}&
 \includegraphics[width = .11\textwidth]{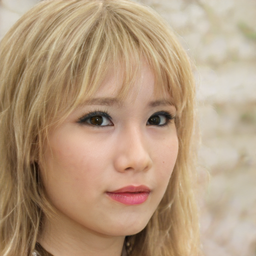}
 &\includegraphics[width = .11\textwidth]{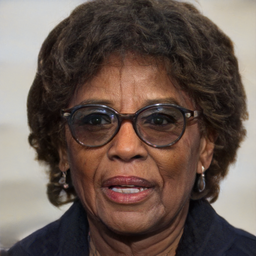}&
 \includegraphics[width = .11\textwidth]{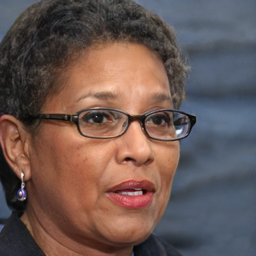}&
 \includegraphics[width = .11\textwidth]{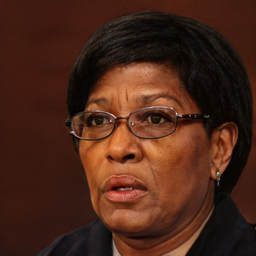}\\
\end{tabular}
\caption{Qualitative comparison with other image manipulation (m) and generation (g) baselines. Many of the other methods appear to ignore the descriptions. StyleCLIP(g) (experimental mode in the codebase, not part of the paper itself) is much more accurate than other methods, but image quality and diversity are lacking. Images generated with Ours are of high visual quality, closely follow the text and are more diverse than other methods.} %
\vspace{-0.45cm}
\label{fig:qualitative_comparison_new}
\end{figure*}
\myparagraph{Baselines.} 
We compare our approach with TediGAN~\cite{xia2021tedigan} and  StyleCLIP~\cite{Patashnik_2021_ICCV} using their official publicly released implementation. We select these two methods because they do not require text data during training, hence a fair comparision with our approach. 
We briefly review each of the methods. 

StyleCLIP provides three ways for doing image generation/manipulation: Optimization, Mapper and Global directions. We compare to the Optimization method because only this method works for open-world captions without re-training or instance-level hyperparameter selection. The Optimization method uses CLIP to guide the iterative modifications of images in StyleGAN's latent space. StyleCLIP Optimization provides two modes in their official Colab notebook\footnote{Colab notebook available at \url{http://colab.research.google.com/github/orpatashnik/StyleCLIP/blob/main/notebooks/optimization_playground.ipynb}}: {\it free generation} and {\it edit}. The ``free generation" mode is not mentioned in the paper as StyleCLIP focuses on %
editing. We include this mode, as it is the closest to our task and achieves the best retrieval accuracy among all baselines.
\begin{itemize}[topsep=0pt,itemsep=-1ex,partopsep=1ex,parsep=1ex, leftmargin=*]

\item {\noindent \it StyleCLIP (g).} The free generation (g) mode can be found in the drop-down menu of ``\text{experiment\_mode}" in the Colab notebook. It initializes the latent vector with the mean latent vector of StyleGAN's latent space and gradually minimizes the CLIP loss between the output and the text. We turn ``stylespace" flag on as it provides best possible visual quality. %

\item {\noindent \it StyleCLIP (m).} The other mode of StyleCLIP Optimization is image manipulation (m) which first generates a face using StyleGANv2 and then manipulates this specific image according to the caption. We use the default hyperparameters in the GitHub code base.
\end{itemize}
TediGAN also does instance-level optimization according to given text prompt, providing two modes:
\begin{itemize}[topsep=0pt,itemsep=-1ex,partopsep=1ex,parsep=1ex, leftmargin=*]
\item {\noindent \it TediGAN (g).} This is the free generation (g) mode of TediGAN. TediGAN contains a text encoder that is trained to project text to the latent space of StyleGAN. It first samples a latent code in StyleGAN's latent space and maps the text to the same latent space with the text encoder. Certain attributes are altered while minimizing the distance between the image and text latent vectors.

\item {\noindent \it TediGAN (m). } This is the image manipulation (m) mode of TediGAN. It requires an input image to start with. We use StyleGANv2 to generate random faces as the inputs to TediGAN (m). A GAN inversion model is used to obtain the latent code corresponding to the input image. The rest is the same as for TediGAN (g).
\end{itemize}
{\noindent We also consider two variants of our method, namely,}
\begin{itemize}[topsep=0pt,itemsep=-1ex,partopsep=1ex,parsep=1ex,leftmargin=*]
    \item {\noindent \it Ours (Pt).} This variant do not use the proposed non-parametric sampling scheme and directly uses a \textit{point (Pt)} estimate, \ie, CLIP's {\it text} embedding is directly passed to the decoder which was trained on {\it image} embeddings.
    \item {\noindent \it Ours (Img).} This variant samples $\vz$ from the encoder's output distribution conditioned on another image, instead of a standard Gaussian. The image can be manually selected by the user or randomly selected to provide extra guidance for details that are not included in the text prompt (\eg, commonly the pose). %
\end{itemize}
\begin{figure}[t]
\centering
\setlength{\tabcolsep}{1pt}
\setlength\arrayrulewidth{1.5pt}
\begin{tabular}{c|ccc}%
Input &\multicolumn{3}{c}{Output} \\
\begin{subfigure}{0.115\textwidth}\centering\includegraphics[width = \textwidth]{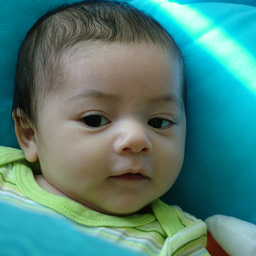}\end{subfigure}&
\begin{subfigure}{0.115\textwidth}\centering\includegraphics[width = \textwidth]{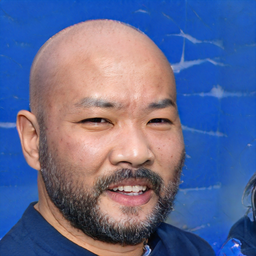}\end{subfigure}&
\begin{subfigure}{0.115\textwidth}\centering\includegraphics[width = \textwidth]{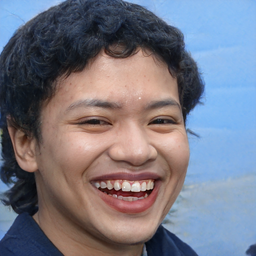}\end{subfigure}&
\begin{subfigure}{0.115\textwidth}\centering\includegraphics[width = \textwidth]{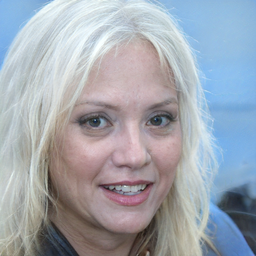}\end{subfigure}\\

\begin{subfigure}{0.115\textwidth}\centering\includegraphics[width = \textwidth]{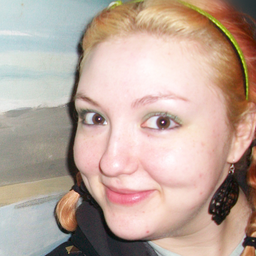}\end{subfigure}&
\begin{subfigure}{0.115\textwidth}\centering\includegraphics[width = \textwidth]{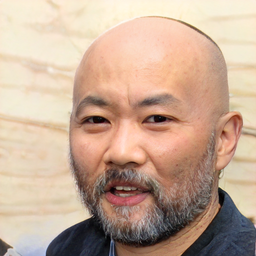}\end{subfigure}&
\begin{subfigure}{0.115\textwidth}\centering\includegraphics[width = \textwidth]{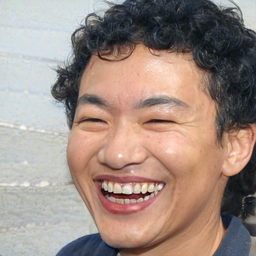}\end{subfigure}&
\begin{subfigure}{0.115\textwidth}\centering\includegraphics[width = \textwidth]{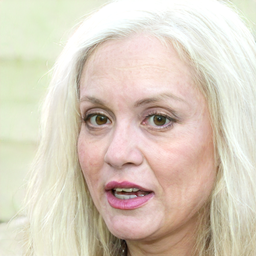}\end{subfigure} %
\end{tabular}
\vspace{0.02cm}
\caption{Results of optional image guidance. The left-most column shows images used as input to the CVAE, while the CLIP embedding comes from a different prompt (see Appendix for prompts used) for each of the rest of the columns. We find the pose and the background color are usually captured by the variational encoder. }
\vspace{-0.3cm}
\label{fig:pose_compare}
\end{figure}
\myparagraph{Implementation details.} We use the publicly available ``ViT-B/16" CLIP model from OpenAI\footnote{Pre-trained model availiable at \url{github.com/openai/CLIP}}. Similarly, StyleGANv2 generator pre-trained on FFHQ has also been released. In our model, the bottle-neck dimension (dimension of $\vz$) is 128 while dimension of $\ve$, in~\equref{eq:mlpe}, is 512. We train our method for 0.5M iterations, using a learning rate of $2e^{-4}$ with weights of $1.0, 1.0, 2e^{-4}~\text{and}~ 0.2$ on $\mathcal{L}_{\text{LPIPS}}$, $\mathcal{L}_{\text{CLIP cycle}}$, $\mathcal{L}_{\vw \text{-norm}}$ and $\mathcal{L}_{\text{KL-divergence}}$ respectively.

\subsection{Quantitative Results}
Quantitative results are reported in~\tabref{tab: results}. As can be seen, Ours achieves the highest retrieval accuracy out of all the methods, \ie, the generated images more closely matches the given text prompt. StyleCLIP (g) also achieved high retrieval accuracy, but with much lower diversity. Next, StyleCLIP (m), TediGan (g) and TediGAN (m) mostly generated images that do not follow the prompt. As expected they achieve the highest diversity as they essentially perform unconditional generation. We compute {\it ID Div.} among random unconditional StyleGANv2 generations and get 0.961. The fact that some of the baselines have {\it ID Div.} values close to that is a support for our claim. We'll discuss this point in more details in \secref{sec:quali}. When comparing among our methods, we observe that using our full model achieves the best result demonstrating the effectiveness of the proposed nonparametric sampling scheme. 

We also report the time for sampling a single image on a single %
QUADRO RTX 6000 GPU. Our approach only requires a single forward pass through the network, and thus is approximately two orders of magnitude faster than prior, iterative optimization based work.

\begin{figure}[t]
\centering
\setlength{\tabcolsep}{0pt}
\begin{tabular}{ccc@{\hspace{0.3em}}ccc}

\multicolumn{6}{c}{\small A photo of a Black woman with curly hair.} \\
\includegraphics[width = .08\textwidth]{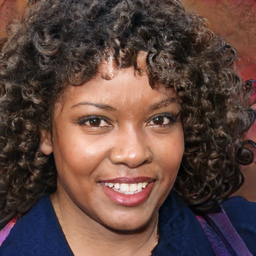}&
 \includegraphics[width = .08\textwidth]{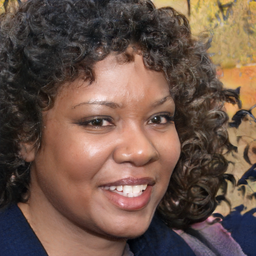}&
\includegraphics[width = .08\textwidth]{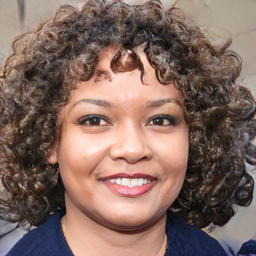}&
 \includegraphics[width = .08\textwidth]{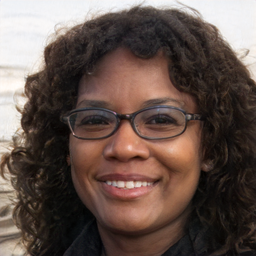}&
 \includegraphics[width = .08\textwidth]{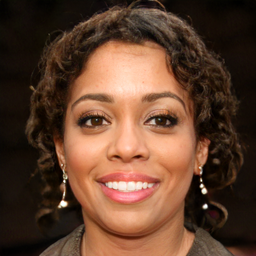}&
\includegraphics[width = .08\textwidth]{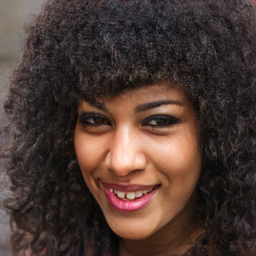}\\

\multicolumn{6}{c}{\small A photo of a smiling small child with blond hair.} \\
\includegraphics[width = .08\textwidth]{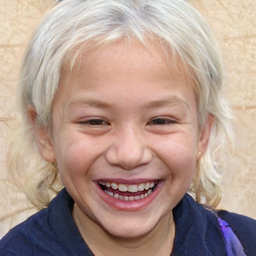}&
 \includegraphics[width = .08\textwidth]{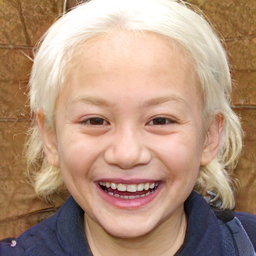}&
 \includegraphics[width = .08\textwidth]{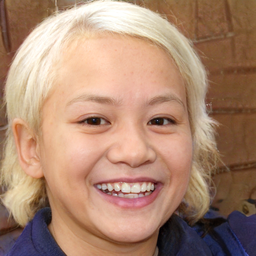}&
\includegraphics[width = .08\textwidth]{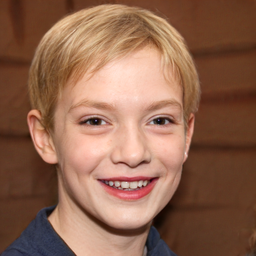}&
 \includegraphics[width = .08\textwidth]{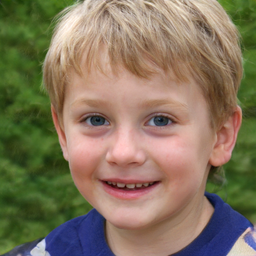}&
 \includegraphics[width = .08\textwidth]{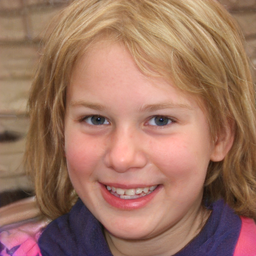}\\

\multicolumn{6}{c}{\small A young pale girl with long dark hair and acne.} \\
\includegraphics[width = .08\textwidth]{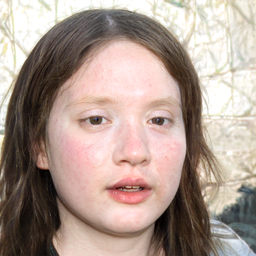}&
 \includegraphics[width = .08\textwidth]{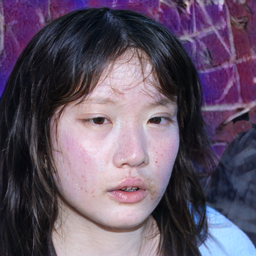}&
 \includegraphics[width = .08\textwidth]{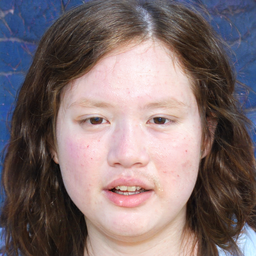}&
\includegraphics[width = .08\textwidth]{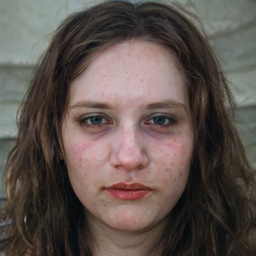}&
 \includegraphics[width = .08\textwidth]{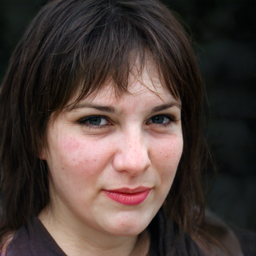}&
 \includegraphics[width = .08\textwidth]{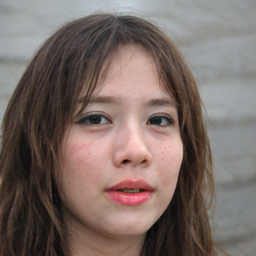}\\

\multicolumn{3}{c}{\small Ours (Pt)} & \multicolumn{3}{c}{\small Ours}  
\end{tabular}
\caption{Comparison of Ours (Pt) and Ours. The non-parametric conditional sampling in Ours significantly improves diversity and, often, visual realism.
}
\vspace{-0.4cm}
\label{fig:gaussian_knn_compare_new}
\end{figure}

\subsection{Qualitative Results}\label{sec:quali}
In~\figref{fig:qualitative_comparison_new}, we qualitatively compare our proposed approach with TediGAN and StyleCLIP. We encourage readers to zoom-in to around 500\% scale to better inspect the details. Ours provides images that are consistent with the text, diverse and with high visual quality. While StyleCLIP (g) captures a large portion of the content in text prompts, its visual quality is poor and lacks diversity among images for given text. On the other hand, StyleCLIP (m) produces highly diverse results, but often the images are unrelated to the given caption. Its performance depends highly on the original face it starts with. For example, the first example of prompt (b) is satisfying because it is initialized with a woman with light-color hair. When the initial image is too ``far away" from the caption, StyleCLIP (m) struggles and often only changes one attribute;\Eg, grey hair in the third image of prompt (a) or blond hair in the third image of prompt (b). We have confirmed with StyleCLIP’s authors that the visual quality for generation is as expected.

TediGAN tends to generates outputs that are unrelated to the captions except in very rare cases. Similar to StyleCLIP, for optimization based methods, it depends heavily on the initializing image. It is more sensitive to certain attributes such as ``old". For example, TediGAN (g) adds wrinkles to the first output of prompt (a) and TediGAN (m) does similar to the first output of prompt (c). Others~\cite{StyleT2I,anyface} also report similar unsatisfactory performance of TediGAN.

In~\figref{fig:pose_compare}, we illustrate the use of image guidance (Ours (Img)) for controlling the pose and background of the image generation, beyond the text prompt. Each column is conditioned with a different text prompt and given the image guidance of the image on the very left. The background and pose of the person follow the guidance image. This is consistent with our hypothesis that $\vz$, and not $\vc$, captures background and pose information.

In~\figref{fig:gaussian_knn_compare_new}, we show images generated by Ours (Pt) and Ours for the same prompts. Ours, using our non-parametric conditional sampling for context, is obviously superior in diversity, providing wider coverage for attributes unspecified in the prompt. It also generates more realistic results than Ours (Pt) (\eg, the hair in the 1st row, acne in the 3rd row). Additionally, Ours is also more accurate,~\eg, the generated images of ``small child" while the outputs from Ours (Pt) look more like teenagers.

\subsection{Ablation Study}
We conduct additional ablation studies on the proposed CVAE architecture and report quantitative result in~\tabref{tab: ablation}. As our proposed encoder takes in both CLIP embedding and image, we consider ablations where the encoder only takes in an image (img only) or a CLIP embedding (CLIP only). We also consider an ablation where the model is a deterministic mapping from CLIP embedding to the image,~\ie, non-variational. Overall, we observe that the full model Ours achieves the highest diversity and best retrieval accuracies. As expected, the non-varitional method performs the worst in diversity. These results demonstrate the efficacy of our CVAE formulation. Models are trained for 0.1M iterations.

\begin{table}[t]
\centering
\setlength{\tabcolsep}{1pt}
\small
\resizebox{\linewidth}{!}{
\begin{tabular}{cccccc}
\specialrule{.15em}{.05em}{.05em}
\multicolumn{1}{c}{}
 & \multicolumn{4}{c}{Correctness (retrieval accuracy)}
 & \multicolumn{1}{c}{Diversity}  \\
\cmidrule(lr){2-5}%
\cmidrule(lr){6-6}
 Models %
 & Top 1$\uparrow$      & Top 5$\uparrow$     & Top 10$\uparrow$     & Top 20$\uparrow$ & ID Div.$\uparrow$     \\
\hline
\hline
Ours        %
& 0.634           & 0.929          & 0.983           & 0.998       & 0.488            \\
Encoder (img only)        %
& 0.314       & 0.725      & 0.890       & 0.980    & 0.090     \\
Encoder (CLIP only)        %
& 0.286       & 0.559      & 0.699       & 0.850   & 0.341     \\
Non-variational        %
& 0.535    & 0.845     & 0.940       & 0.993     & 0.035    \\
\specialrule{.15em}{.05em}{.05em}
\end{tabular}
}
\caption{Ablation study: quantitative metrics of different architecture designs. We considered removing important components of our model and compare the performances.
}
\vspace{-0.4cm}
\label{tab: ablation}
\end{table}

\section{Limitations %
  and other considerations}\label{sec:limitation}

\begin{figure}[t]
\centering
\setlength{\tabcolsep}{2pt}
\begin{tabular}{ccc} %
\includegraphics[width = .32\linewidth]{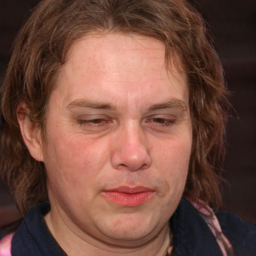}&
 \includegraphics[width = .32\linewidth]{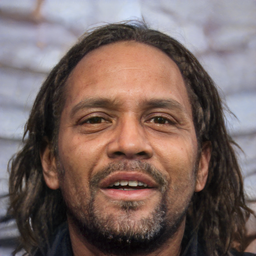}&
\includegraphics[width = .32\linewidth]{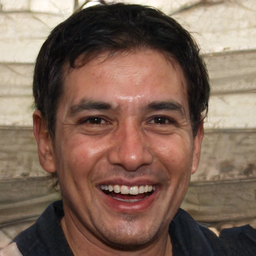}%
\end{tabular}
\caption{Failure cases. From left to right these demonstrate a failure to generate combinations of attributes not present in the original data (``Sad man with closed eyes and lipstick"), generate rare or missing attributes in the original data (``Man with dreadlocks"), and a failure due to the CLIP embedding not capturing negation (``Man showing no teeth")}
\vspace{-0.3cm}
\label{fig:failures}
\end{figure}

{\bf\noindent Limitations.} In \figref{fig:failures}, we show some failure cases. %
The man's eyes in the left image are not fully closed and he does not look sad. This is because there are no exemplars in FFHQ dataset that match the prompt well. In such situation, the user can use Ours (Pt) for better accuracy, while sacrificing some diversity. %
In the middle image, the model struggles to generate dreadlocks because there are limited training data with this attribute. Finally, the model lacks understanding of negation, a known limitation of CLIP.

The last failure case in~\figref{fig:failures} reflects the main limitations of our approach are the limitations of the underlying CLIP embeddings. Despite CLIP's impressive ability to map images and text to the same space, which we leverage in our work, the nature of the cross-modal association in CLIP remains obscure, making some attributes harder to control via text than others. Furthermore, mapping a text description (which could match many potential images) to a single direction in the embedding space is likely insufficient for broader control of image generation, and for other tasks relying on image-text matching; this remains a challenge for further improvement of CLIP-like models.

We have chosen to focus on human faces (``portraits'') because (i) this is an important domain, with clear potential for applications (\eg, synthesis of stock images, movie making, \etc.) and an obvious range of interesting descriptions; and (ii) faces remain by far the domain with the most success in photorealistic image synthesis. Extending our approach to other domains remains subject of future work.

\myparagraph{General concerns.}  There is a potential for mis-use of image synthesis technology, especially if it involves human faces. However, we believe it is less significant for de-novo synthesis (that we do) than for image manipulation that preserves real people's identity (that we do not do). Another concern is the inherent biases present in both the generators and the CLIP embeddings. For instance, we observe that in most cases, when the ethnicity of the target is not specified, the method generates Caucasian faces more often. However, our method offers a way to combat this bias: by explicitly specifying the properties of the target one can ensure that a diverse range of faces can be produced.

\section{Conclusion}\label{sec:conc}
We propose a method for learning an interface between pre-trained text embeddings (CLIP) and a pre-trained image generator (StyleGANv2). Once trained, our interface enables fast conditional image synthesis from natural language. Training our method requires only an unlabeled image set, without associated text information. Our introduced CVAE formulation %
allows modeling diverse visual attributes that are not specified by the text prompt. We also proposed a test time procedure scheme based on non-parametric sampling, which improves quality and robustness of the generation process. This is done by modelling a non-parametric distribution of CLIP's image embedding given text embedding. Empirically, we demonstrate that our~\modelname produces high quality images that more faithfully follow the text prompt and is significantly faster in inference speed compared to recent baselines.

{\small
\clearpage
\bibliographystyle{ieee_fullname}
\bibliography{reference}
}

\clearpage
\appendix
\renewcommand{\thetable}{A\arabic{table}}
\setcounter{table}{0}
\setcounter{figure}{0}
\renewcommand{\thetable}{A\arabic{table}}
\renewcommand\thefigure{A\arabic{figure}}

{\centering \Large \textbf{Appendix}}\\

\noindent This appendix is organized as follows:
\begin{itemize}
\item In~\secref{apx:fid}, we explain why FID is not suitable for evaluating conditional image generation tasks and provide evidence to support our claim.

\item In~\secref{apx:more_baselines}, we provide more qualitative comparison results against the baselines.

\item In~\secref{apx:pt}, we provide examples of Ours (Pt) generates images faithful to prompts when Ours fail, due to the reason combinations of certain attributes not present in the original data.

\item In~\secref{apx:knn}, we provide additional results generated by Ours using~\modelname.

\item In~\secref{apx:NN}, we examine the nearest neighbors of our generated samples in FFHQ dataset, showing our model is able to generate novel faces rather than pure memorization or simple hallucination.

\item In~\secref{apx:prompts}, we provide text prompts used to generate output images in~\figref{fig:pose_compare} in the paper.

\end{itemize}

\section{Discussion on FID}\label{apx:fid}
\myparagraph{Definition.} Fréchet Inception Distance (FID), proposed by Heusel~\etal~\cite{heusel2017gans}, is an evaluation metric for image generation model. Given a set of generated images and another set of real images, FID measures the squared Wasserstein metric between the two sets modeled as a multivariate Gaussian distribution with features extracted from Inception-v3~\cite{szegedy2015going} (trained on ImageNet~\cite{deng2009imagenet}). Formally, Heusel \etal~\cite{heusel2017gans} define FID as:
\bea
\norm{\mu_{\tt g} - \mu_{\tt r}}_2^2 + \left(\Sigma_{\tt g}+\Sigma_{\tt r}-2\left(\Sigma_{\tt g}\Sigma_{\tt r}\right)^{\frac{1}{2}}\right)
\eea
where $\mu_{\tt g/r}$ and $\Sigma_{\tt g/r}$ %
denotes the mean and covariance matrix estimate from the generated/real data in the feature space, respectively. To extract features, FID uses the last pooling layer of Inception-v3 pre-trained on ImageNet. Better (lower) FID means the two distributions are closer in feature space.

\begin{figure*}[t]
\centering
\setlength{\tabcolsep}{1pt}
\small
\begin{tabular}{ccc}
&
\begin{tabular}{l}
    Generated Images ({Input Text: A photo of a smiling girl with black hair).}
\end{tabular}
& FID$\downarrow$\\

\begin{tabular}{>{\begin{sideways}}c<{\end{sideways}}}
{\small StyleGANv2}
\end{tabular}
&
\begin{tabular}{cccccccc}
\includegraphics[width = .11\textwidth]{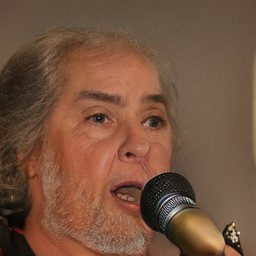}%
\includegraphics[width = .11\textwidth]{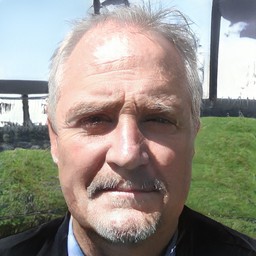}%
\includegraphics[width = .11\textwidth]{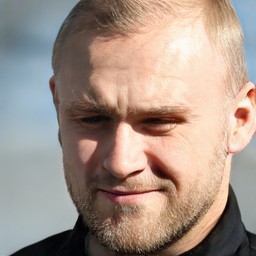}%
\includegraphics[width = .11\textwidth]{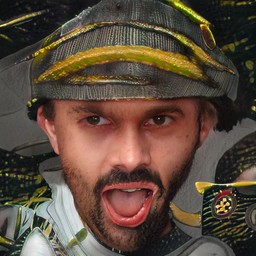}%
\includegraphics[width = .11\textwidth]{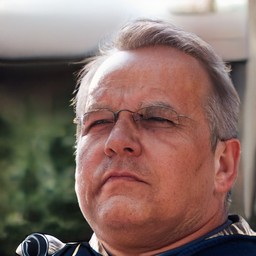}%
\includegraphics[width = .11\textwidth]{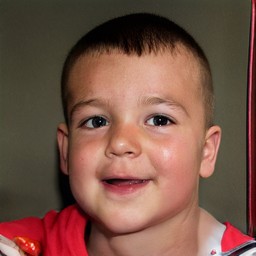}%
\includegraphics[width = .11\textwidth]{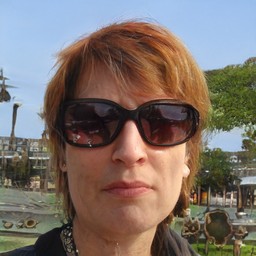}%
\includegraphics[width = .11\textwidth]{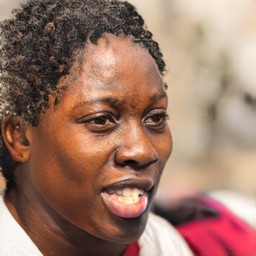}
\end{tabular}
&
\begin{tabular}{c}
4.20
\end{tabular}\\

\begin{tabular}{>{\begin{sideways}}c<{\end{sideways}}}
{\small StyleCLIP (m)}
\end{tabular}
&
\begin{tabular}{cccccccc}
\includegraphics[width = .11\textwidth]{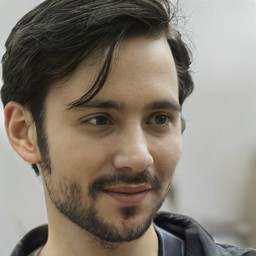}%
\includegraphics[width = .11\textwidth]{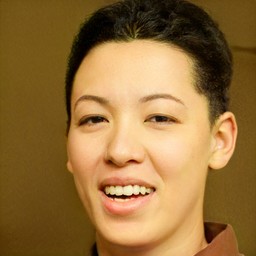}%
\includegraphics[width = .11\textwidth]{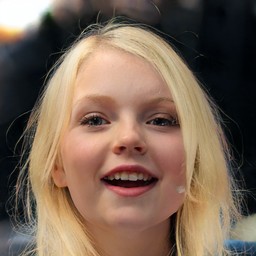}%
\includegraphics[width = .11\textwidth]{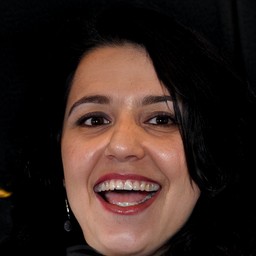}%
\includegraphics[width = .11\textwidth]{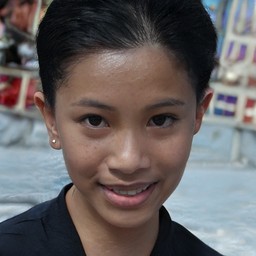}%
\includegraphics[width = .11\textwidth]{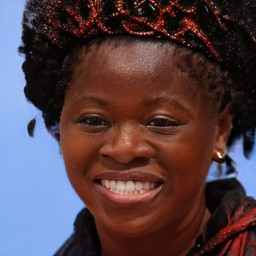}%
\includegraphics[width = .11\textwidth]{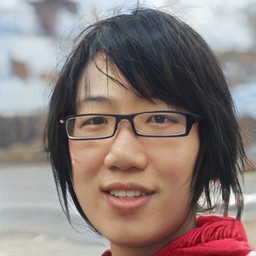}%
\includegraphics[width = .11\textwidth]{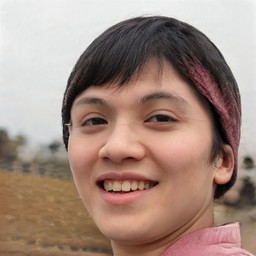}
\end{tabular}
&
\begin{tabular}{c}
5.93
\end{tabular}\\

\begin{tabular}{>{\begin{sideways}}c<{\end{sideways}}}
{\small TediGAN (m)}
\end{tabular}
&
\begin{tabular}{cccccccc}
\includegraphics[width = .11\textwidth]{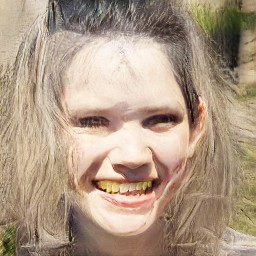}%
\includegraphics[width = .11\textwidth]{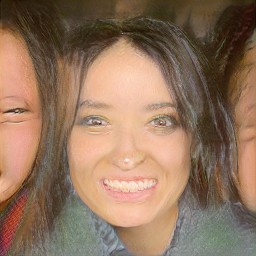}%
\includegraphics[width = .11\textwidth]{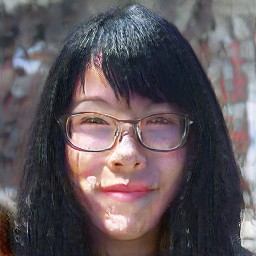}%
\includegraphics[width = .11\textwidth]{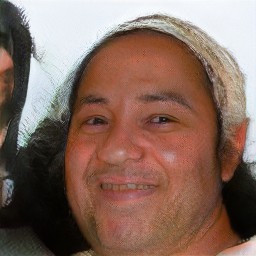}%
\includegraphics[width = .11\textwidth]{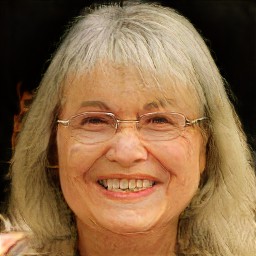}%
\includegraphics[width = .11\textwidth]{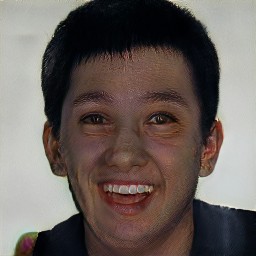}%
\includegraphics[width = .11\textwidth]{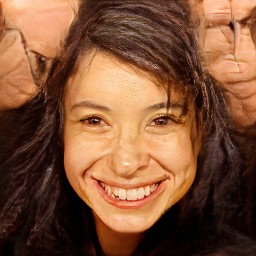}%
\includegraphics[width = .11\textwidth]{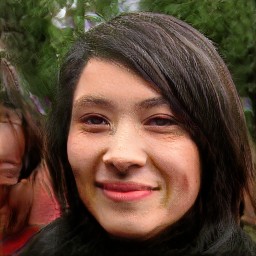}%
\end{tabular}
&
\begin{tabular}{c}
23.38
\end{tabular}\\
 
\begin{tabular}{>{\begin{sideways}}c<{\end{sideways}}}
{\small StyleCLIP (g)}
\end{tabular}
&
\begin{tabular}{cccccccc}
\includegraphics[width = .11\textwidth]{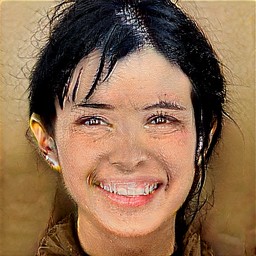}%
\includegraphics[width = .11\textwidth]{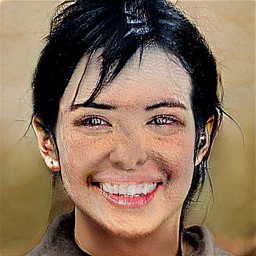}%
\includegraphics[width = .11\textwidth]{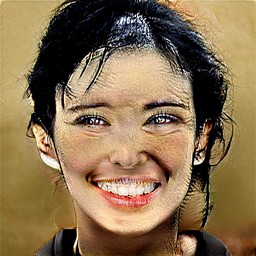}%
\includegraphics[width = .11\textwidth]{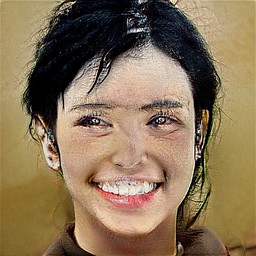}%
\includegraphics[width = .11\textwidth]{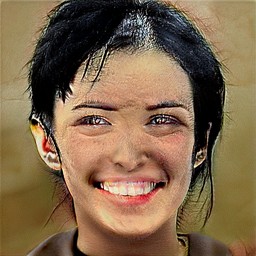}%
\includegraphics[width = .11\textwidth]{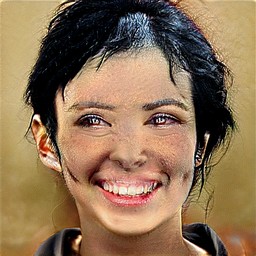}%
\includegraphics[width = .11\textwidth]{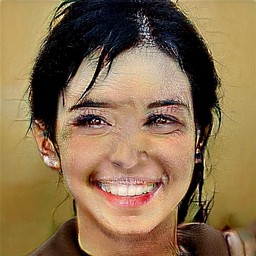}%
\includegraphics[width = .11\textwidth]{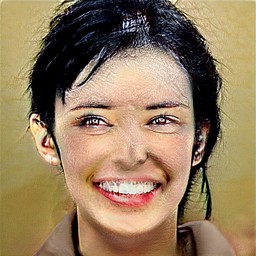}%
\end{tabular}
&
\begin{tabular}{c}
71.74
\end{tabular}\\

\begin{tabular}{>{\begin{sideways}}c<{\end{sideways}}}
{\small TediGAN (g)}
\end{tabular}
&
\begin{tabular}{cccccccc}
\includegraphics[width = .11\textwidth]{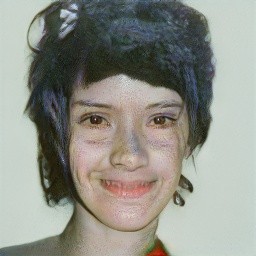}%
\includegraphics[width = .11\textwidth]{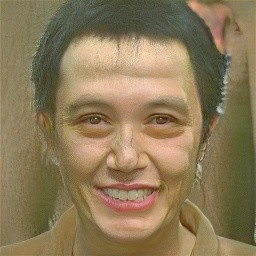}%
\includegraphics[width = .11\textwidth]{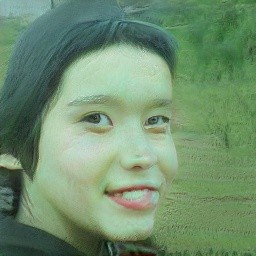}%
\includegraphics[width = .11\textwidth]{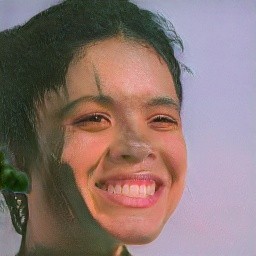}%
\includegraphics[width = .11\textwidth]{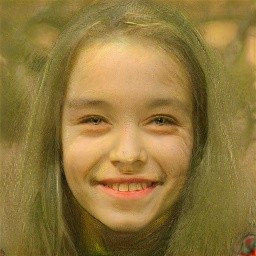}%
\includegraphics[width = .11\textwidth]{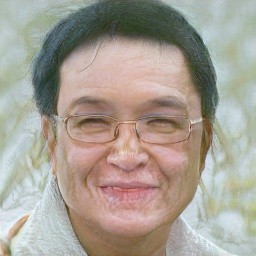}%
\includegraphics[width = .11\textwidth]{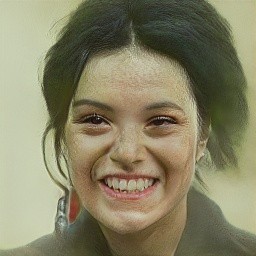}%
\includegraphics[width = .11\textwidth]{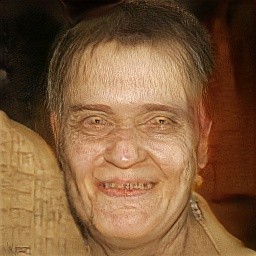}%
\end{tabular}
&
\begin{tabular}{c}
32.51
\end{tabular}\\

\begin{tabular}{>{\begin{sideways}}c<{\end{sideways}}}
{\small Ours}
\end{tabular}
&
\begin{tabular}{cccccccc}
\includegraphics[width = .11\textwidth]{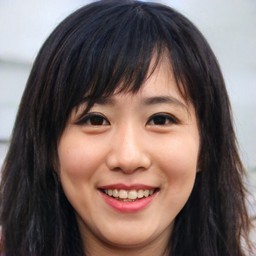}%
\includegraphics[width = .11\textwidth]{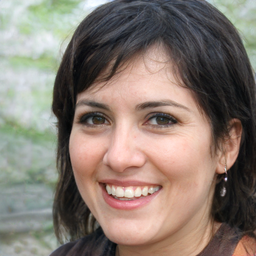}%
\includegraphics[width = .11\textwidth]{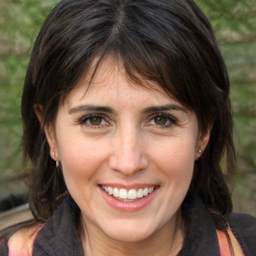}%
\includegraphics[width = .11\textwidth]{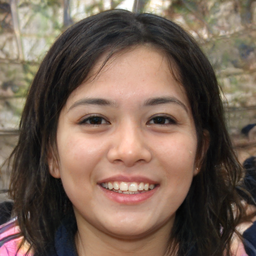}%
\includegraphics[width = .11\textwidth]{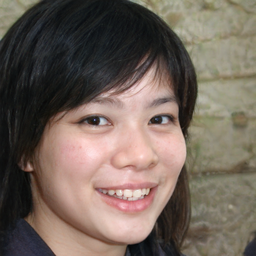}%
\includegraphics[width = .11\textwidth]{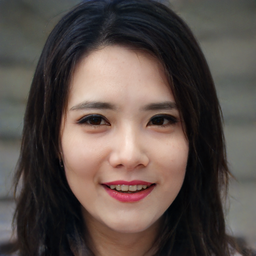}%
\includegraphics[width = .11\textwidth]{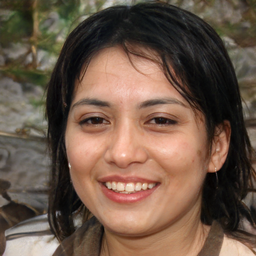}%
\includegraphics[width = .11\textwidth]{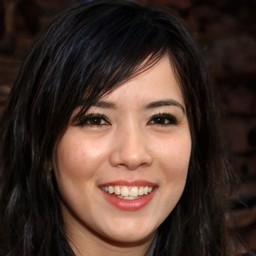}%
\end{tabular}
&
\begin{tabular}{c}
60.72
\end{tabular}\\

\end{tabular}
\caption{FID and sample images from the same caption given on top. The eight images for each method are uncurated. In the first row, we include FID and random unconditional outputs from the vanilla StyleGANv2 for reference. As can be seen, FID cannot correctly reflect performance of models for conditional generation.} 
\label{fig:supp_fid}
\end{figure*}

\myparagraph{FID is not suitable for conditional generation.} 
The FFHQ data is unconditional, \ie, it does not come with text captions. On the other hand, the generated images from the models are conditioned on the 400 text prompts. In other words, unless the text covers the entire possible faces of FFHQ, the two distributions are simply different distributions. Hence, FID is not a suitable evaluation metric, as following the text prompts could potentially penalize the metric. We illustrate this point in~\figref{fig:supp_fid}, where we show images generated  (for a single prompt) and corresponding FID, from the unconditional StyleGANv2, baselines and ours. The result reveals the shortcomings of FID:  (a) approaches that simply ignores the text achieves low FID and (b) FID does not necessary correlates with the perceived image quality.

\section{More qualitative comparisons with baselines}\label{apx:more_baselines}
\begin{figure*}[t]
\centering
\setlength{\tabcolsep}{0pt}
\begin{tabular}{
>{\begin{sideways}}c<{\end{sideways}}
ccc@{\hspace{0.3em}}ccc@{\hspace{0.3em}}ccc}

& \multicolumn{3}{c}{ \small 
    \begin{tabular}{c}
    (a) An Asian girl with purple hair.
  \end{tabular}
  } &\multicolumn{3}{c}{ \small 
  \begin{tabular}{c}
   (b) A white person with blue eyes \\ and red hair slightly opens mouth.
  \end{tabular}
  } &\multicolumn{3}{C{.33\textwidth}}{ \small 
  \begin{tabular}{c}
  (c) A middle-aged Black woman \\ with heavy makeup.
  \end{tabular}
  } \\

{\small StyleCLIP (m)} & \includegraphics[width = .11\textwidth]{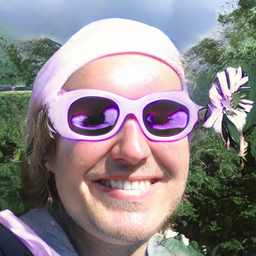}&
 \includegraphics[width = .11\textwidth]{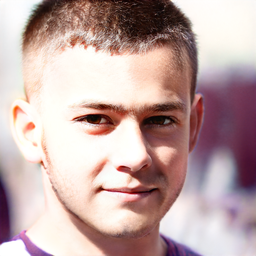}&
 \includegraphics[width = .11\textwidth]{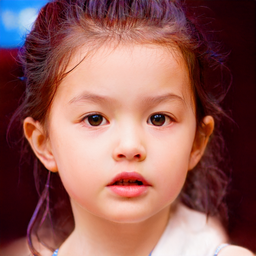}&
\includegraphics[width = .11\textwidth]{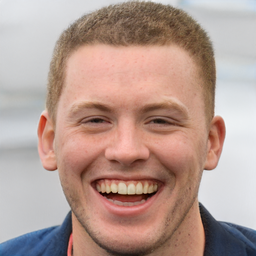}&
 \includegraphics[width = .11\textwidth]{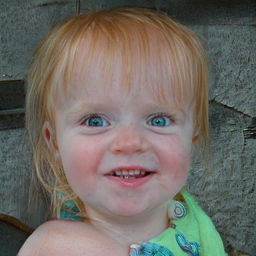}&
 \includegraphics[width = .11\textwidth]{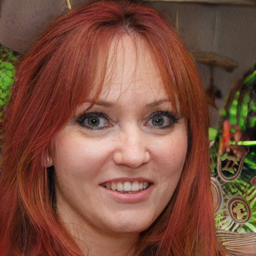}
 &\includegraphics[width = .11\textwidth]{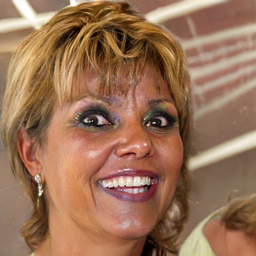}&
 \includegraphics[width = .11\textwidth]{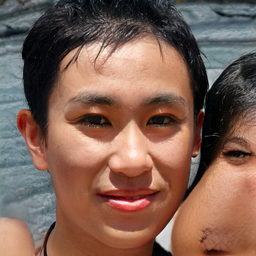}&
 \includegraphics[width = .11\textwidth]{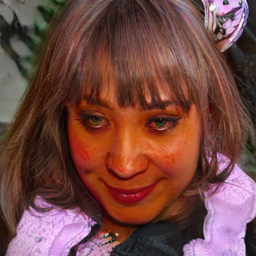}\\
 
{\small TediGAN (m)} & \includegraphics[width = .11\textwidth]{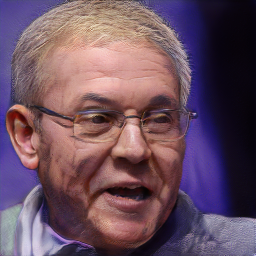}&
 \includegraphics[width = .11\textwidth]{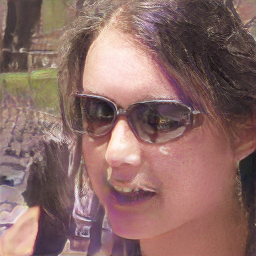}&
 \includegraphics[width = .11\textwidth]{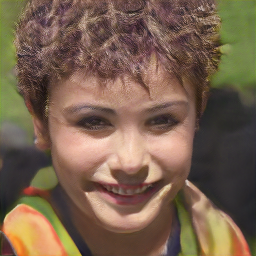}&
\includegraphics[width = .11\textwidth]{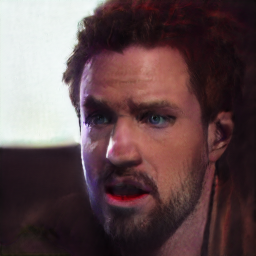}&
 \includegraphics[width = .11\textwidth]{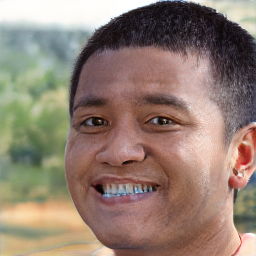}&
 \includegraphics[width = .11\textwidth]{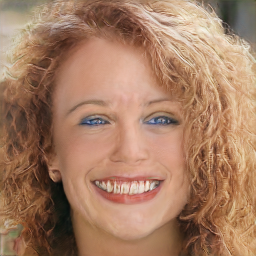}&
 \includegraphics[width = .11\textwidth]{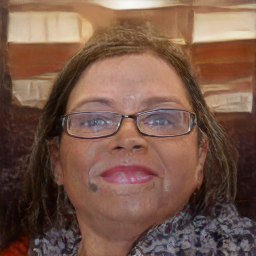}&
 \includegraphics[width = .11\textwidth]{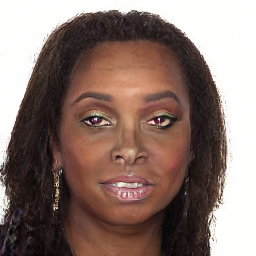}&
 \includegraphics[width = .11\textwidth]{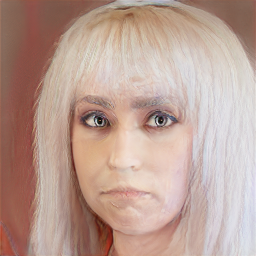} \\
{\small StyleCLIP (g) }& \includegraphics[width = .11\textwidth]{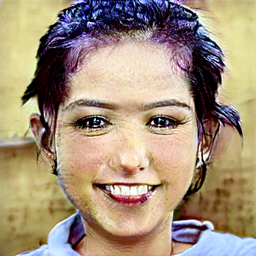}&
 \includegraphics[width = .11\textwidth]{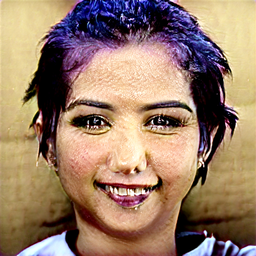}&
 \includegraphics[width = .11\textwidth]{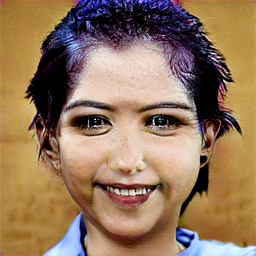}&
\includegraphics[width = .11\textwidth]{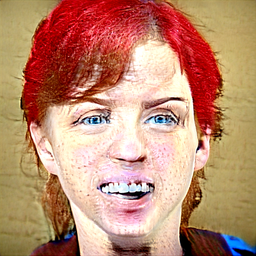}&
 \includegraphics[width = .11\textwidth]{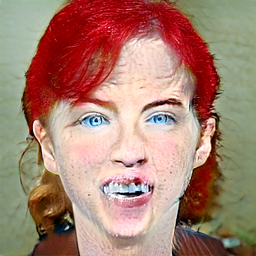}&
 \includegraphics[width = .11\textwidth]{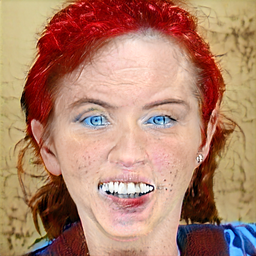}
 &\includegraphics[width = .11\textwidth]{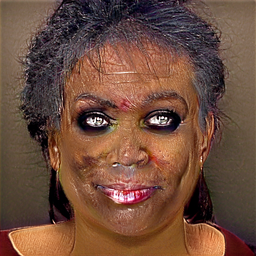}&
 \includegraphics[width = .11\textwidth]{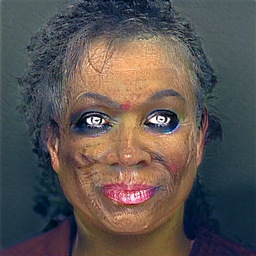}&
 \includegraphics[width = .11\textwidth]{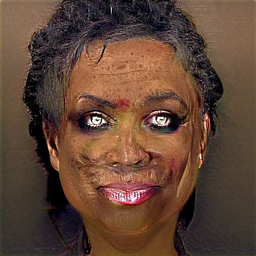}\\
{\small TediGAN (g)} & \includegraphics[width = .11\textwidth]{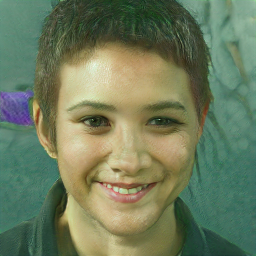}&
 \includegraphics[width = .11\textwidth]{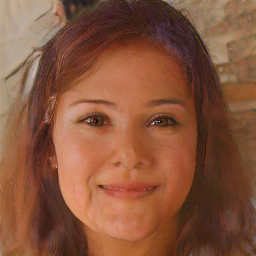}&
 \includegraphics[width = .11\textwidth]{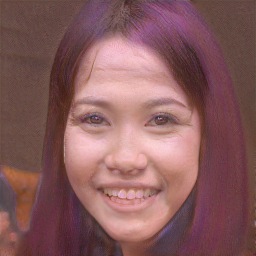}&
\includegraphics[width = .11\textwidth]{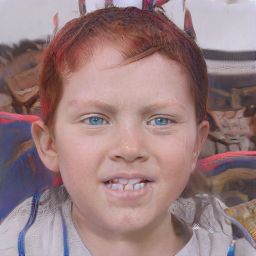}&
 \includegraphics[width = .11\textwidth]{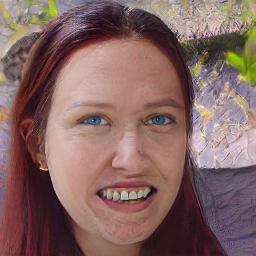}&
 \includegraphics[width = .11\textwidth]{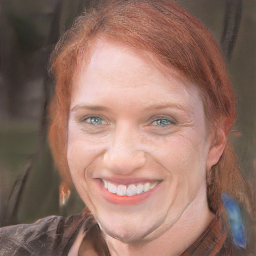}&
 \includegraphics[width = .11\textwidth]{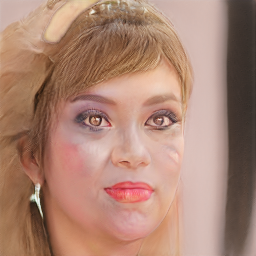}&
 \includegraphics[width = .11\textwidth]{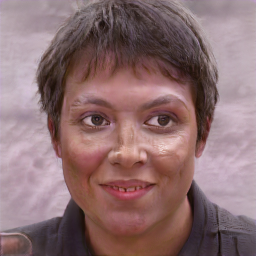}&
 \includegraphics[width = .11\textwidth]{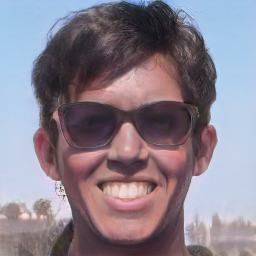} \\

\hspace{0.55cm}\textbf{Ours} & \includegraphics[width = .11\textwidth]{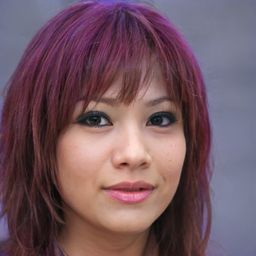}&
 \includegraphics[width = .11\textwidth]{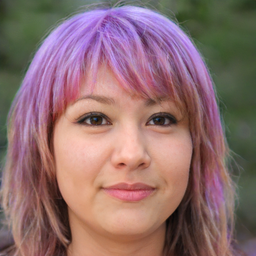}&
 \includegraphics[width = .11\textwidth]{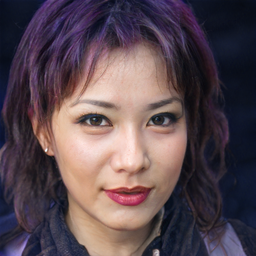}&
\includegraphics[width = .11\textwidth]{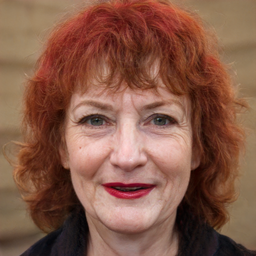}&
 \includegraphics[width = .11\textwidth]{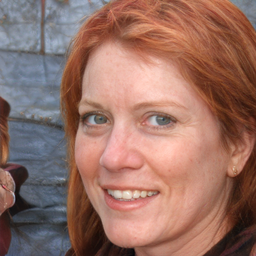}&
 \includegraphics[width = .11\textwidth]{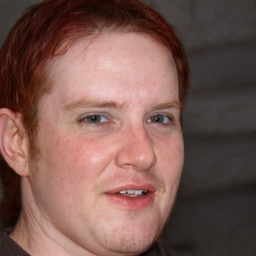}&
 \includegraphics[width = .11\textwidth]{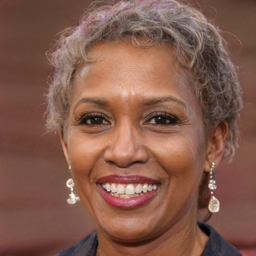}&
 \includegraphics[width = .11\textwidth]{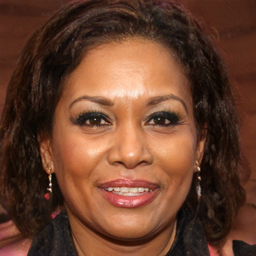}&
 \includegraphics[width = .11\textwidth]{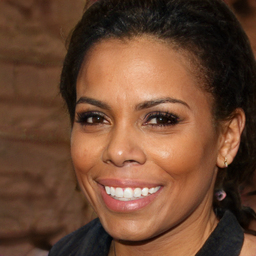}\\
\end{tabular}
\caption{More qualitative comparison with image manipulation (m) and generation (g) baselines.} %
\label{fig:supp_baselines}
\end{figure*}

In~\figref{fig:supp_baselines}, we provide additional visualization of our approach compared to baselines. 

\section{Ours (Pt) vs. Ours for out-of-distribution prompts}\label{apx:pt}
\begin{figure*}[t]
\centering
\setlength{\tabcolsep}{0pt}
\begin{tabular}{ccc@{\hspace{0.5em}}ccc}

\multicolumn{6}{c}{\small 
    \begin{tabular}{c}
    Sad man with closed eyes and lipstick.
  \end{tabular}} \\
\includegraphics[width = .15\textwidth]{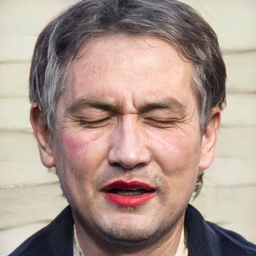}&
 \includegraphics[width = .15\textwidth]{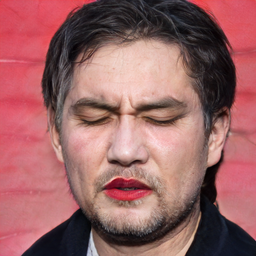}&
\includegraphics[width = .15\textwidth]{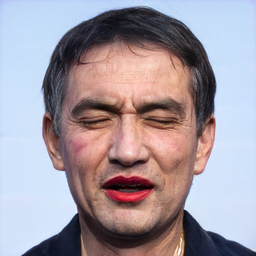}&
 \includegraphics[width = .15\textwidth]{figures/ours_knn/Sad man with closed eyes and lipstick/knn_2.png}&
 \includegraphics[width = .15\textwidth]{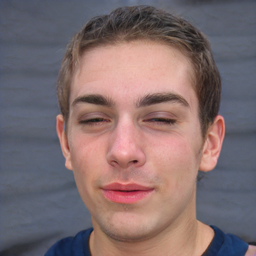}&
\includegraphics[width = .15\textwidth]{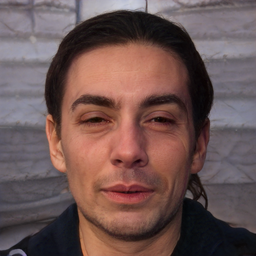}\\

\multicolumn{6}{c}{\small 
    \begin{tabular}{c}
    A female with no eyebrows.
  \end{tabular}} \\
\includegraphics[width = .15\textwidth]{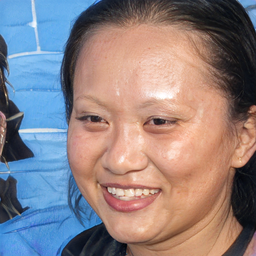}&
 \includegraphics[width = .15\textwidth]{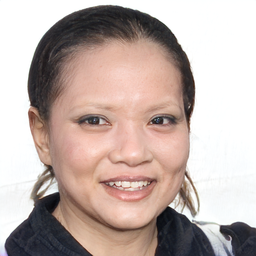}&
 \includegraphics[width = .15\textwidth]{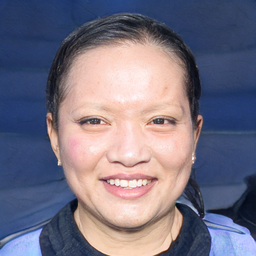}&
\includegraphics[width = .15\textwidth]{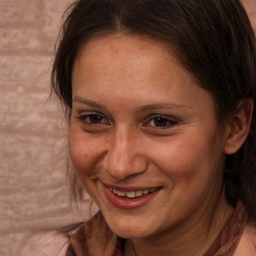}&
 \includegraphics[width = .15\textwidth]{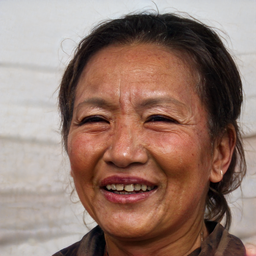}&
 \includegraphics[width = .15\textwidth]{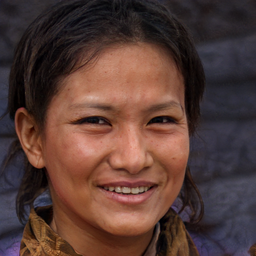}\\

\multicolumn{6}{c}{\small 
    \begin{tabular}{c}
    Kid with large ears and no hair.
  \end{tabular}} \\
\includegraphics[width = .15\textwidth]{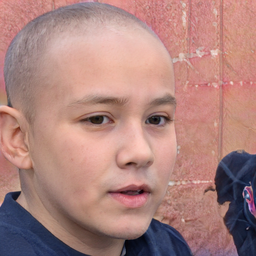}&
 \includegraphics[width = .15\textwidth]{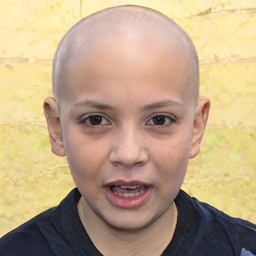}&
 \includegraphics[width = .15\textwidth]{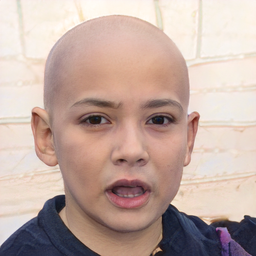}&
\includegraphics[width = .15\textwidth]{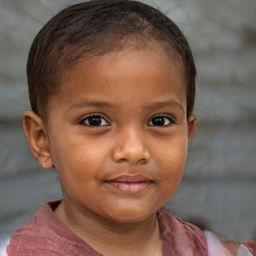}&
 \includegraphics[width = .15\textwidth]{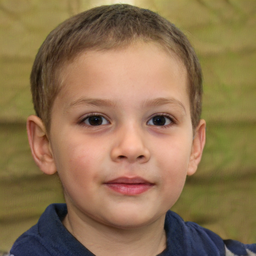}&
 \includegraphics[width = .15\textwidth]{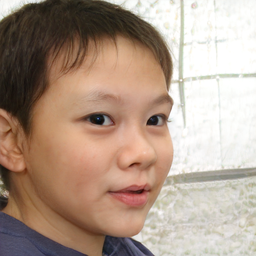}\\

\multicolumn{3}{c}{\small Ours (Pt)} & \multicolumn{3}{c}{\small Ours}  
\end{tabular}
\caption{Ours (Pt) vs. Ours for out-of-distribution prompts. When it is hard to find exemplars in FFHQ that match the out-of-distribution text prompts, Ours (Pt) gives more accurate outputs while sacrificing some diversity.
}
\vspace{-0.4cm}
\label{fig:vs}
\end{figure*}

In~\figref{fig:vs}, we show results on out-of-distribution prompts,\ie, the caption does not resemble any image that is in the FFHQ dataset. In such cases, we observe that Ours (Pt) more acutely follows the prompt but at the cost of diversity.

\section{Additional results generated by Ours}\label{apx:knn}
\begin{figure*}[t]
\centering
\setlength{\tabcolsep}{0pt}
\renewcommand{\arraystretch}{0.0}
\begin{tabular}{cccc}

\multicolumn{4}{c}{\small 
    \begin{tabular}{c}
    A Black female wearing bright lipstick.
  \end{tabular}} \\
\includegraphics[width = .21\textwidth]{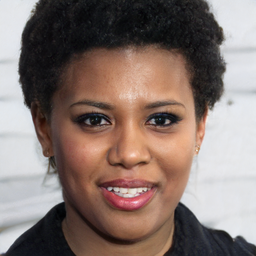}&
\includegraphics[width = .21\textwidth]{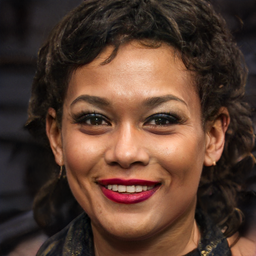}&
\includegraphics[width = .21\textwidth]{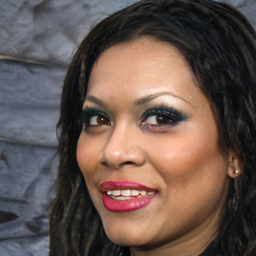}&
\includegraphics[width = .21\textwidth]{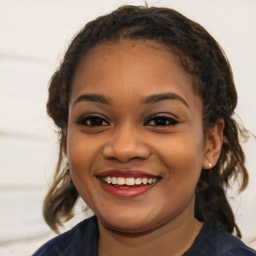}\\
\includegraphics[width = .21\textwidth]{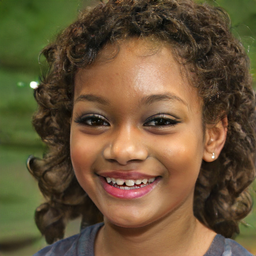}&
\includegraphics[width = .21\textwidth]{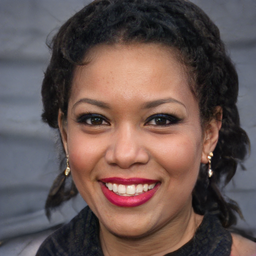}&
\includegraphics[width = .21\textwidth]{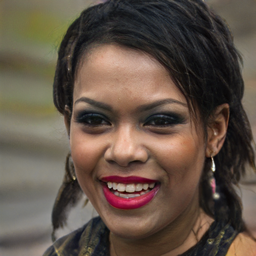}&
\includegraphics[width = .21\textwidth]{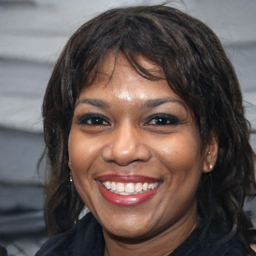} \\ [5pt]

\multicolumn{4}{c}{\small 
    \begin{tabular}{c}
    A chubby person on dark background.
  \end{tabular}} \\
\includegraphics[width = .21\textwidth]{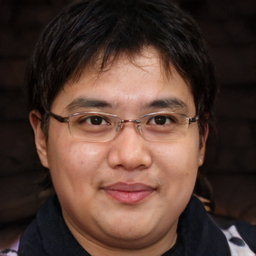}&
\includegraphics[width = .21\textwidth]{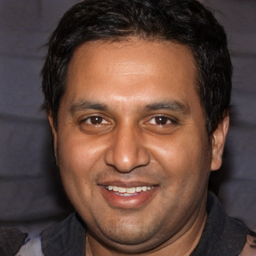}&
\includegraphics[width = .21\textwidth]{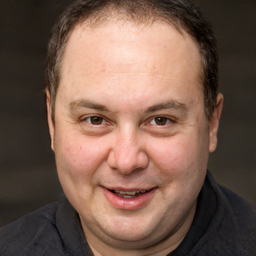}&
\includegraphics[width = .21\textwidth]{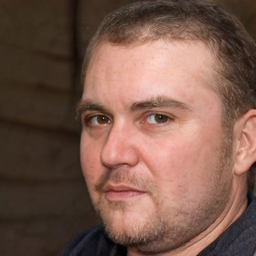}\\
\includegraphics[width = .21\textwidth]{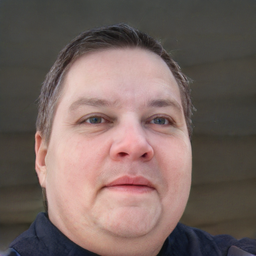}&
\includegraphics[width = .21\textwidth]{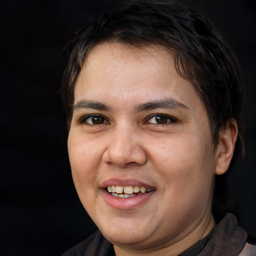}&
\includegraphics[width = .21\textwidth]{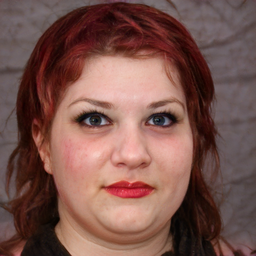}&
\includegraphics[width = .21\textwidth]{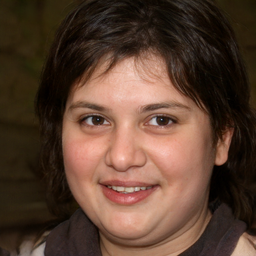}\\ [5pt]

\multicolumn{4}{c}{\small 
    \begin{tabular}{c}
    A smiling woman with curly blond hair wearing sunglasses.
  \end{tabular}} \\
\includegraphics[width = .21\textwidth]{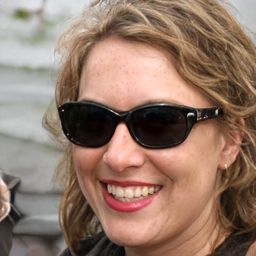}&
\includegraphics[width = .21\textwidth]{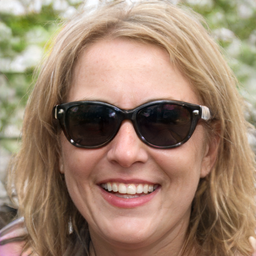}&
\includegraphics[width = .21\textwidth]{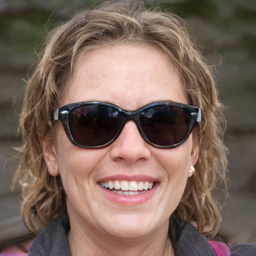}&
\includegraphics[width = .21\textwidth]{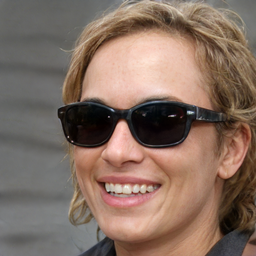}\\
\includegraphics[width = .21\textwidth]{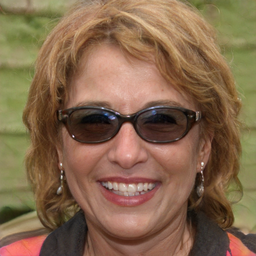}&
\includegraphics[width = .21\textwidth]{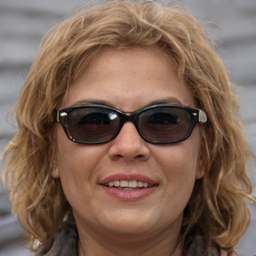}&
\includegraphics[width = .21\textwidth]{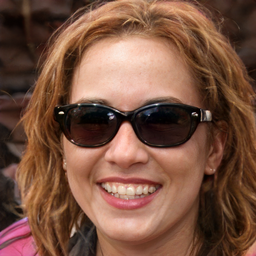}&
\includegraphics[width = .21\textwidth]{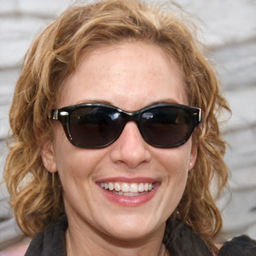}

\end{tabular}
\caption{Additional results for Ours. The text prompt on top is used to generate the eight images below it.
}
\label{fig:extra1}
\end{figure*}

\begin{figure*}[t]
\centering
\setlength{\tabcolsep}{0pt}
\renewcommand{\arraystretch}{0.0}
\begin{tabular}{cccc}

\multicolumn{4}{c}{\small 
    \begin{tabular}{c}
    A calm brunette with pointy nose.
  \end{tabular}} \\
\includegraphics[width = .21\textwidth]{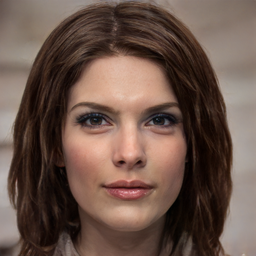}&
\includegraphics[width = .21\textwidth]{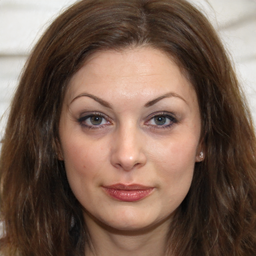}&
\includegraphics[width = .21\textwidth]{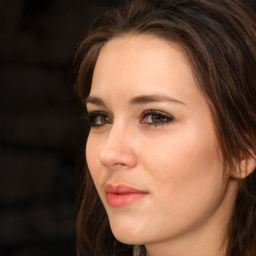}&
\includegraphics[width = .21\textwidth]{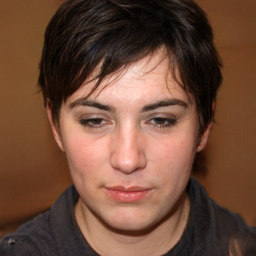}\\
\includegraphics[width = .21\textwidth]{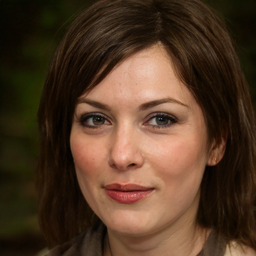}&
\includegraphics[width = .21\textwidth]{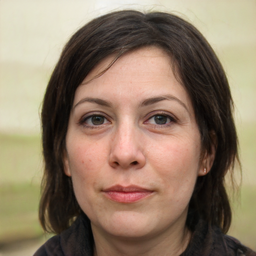}&
\includegraphics[width = .21\textwidth]{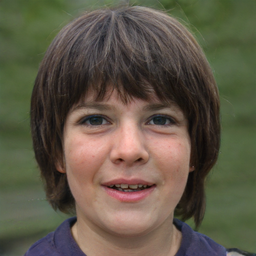}&
\includegraphics[width = .21\textwidth]{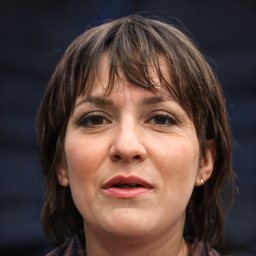}\\ [5pt]

\multicolumn{4}{c}{\small 
    \begin{tabular}{c}
    A photo of a person with big nose.
  \end{tabular}} \\
\includegraphics[width = .21\textwidth]{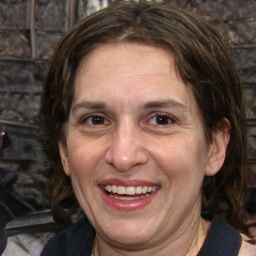}&
\includegraphics[width = .21\textwidth]{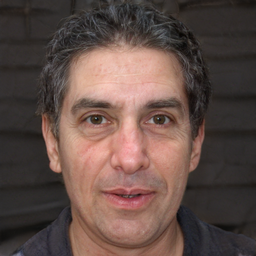}&
\includegraphics[width = .21\textwidth]{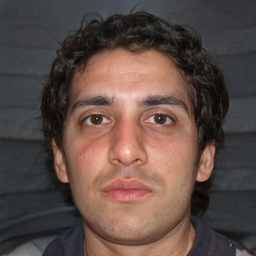}&
\includegraphics[width = .21\textwidth]{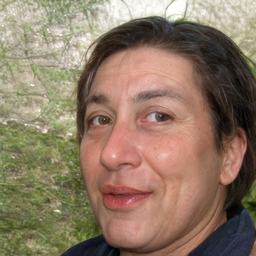}\\
\includegraphics[width = .21\textwidth]{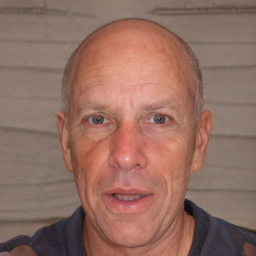}&
\includegraphics[width = .21\textwidth]{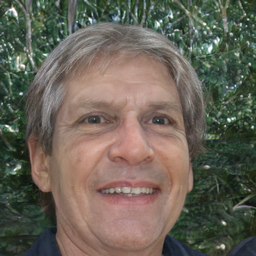}&
\includegraphics[width = .21\textwidth]{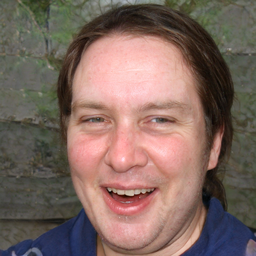}&
\includegraphics[width = .21\textwidth]{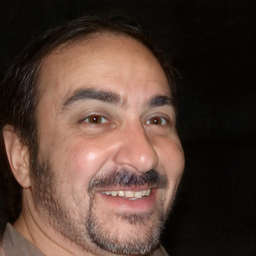}\\ [5pt]

\multicolumn{4}{c}{\small 
    \begin{tabular}{c}
    A curious baby.
  \end{tabular}} \\
\includegraphics[width = .21\textwidth]{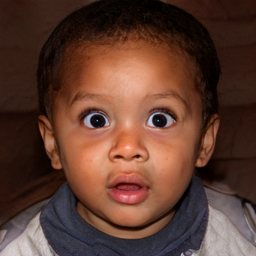}&
\includegraphics[width = .21\textwidth]{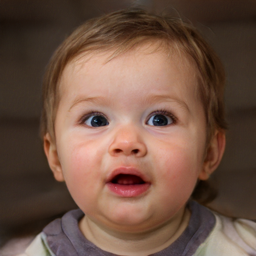}&
\includegraphics[width = .21\textwidth]{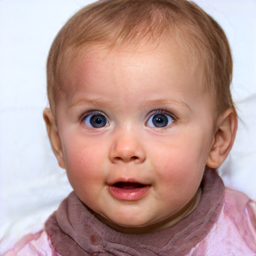}&
\includegraphics[width = .21\textwidth]{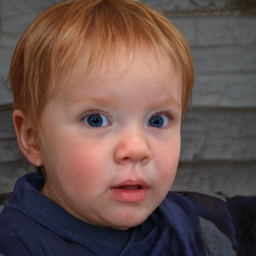}\\
\includegraphics[width = .21\textwidth]{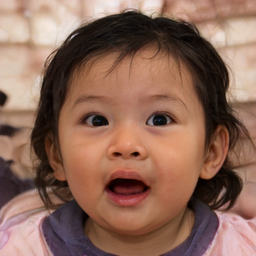}&
\includegraphics[width = .21\textwidth]{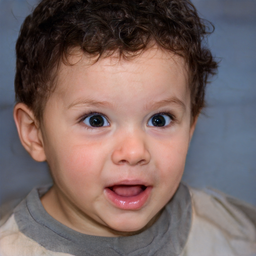}&
\includegraphics[width = .21\textwidth]{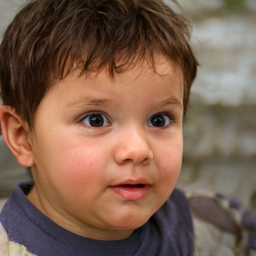}&
\includegraphics[width = .21\textwidth]{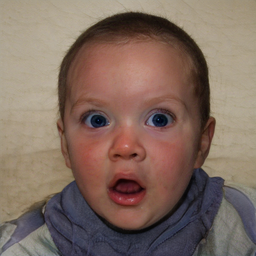}

\end{tabular}
\caption{Additional results for Ours. The text prompt on top is used to generate the eight images below it.
}
\label{fig:extra2}
\end{figure*}

In~\figref{fig:extra1},~\figref{fig:extra2} and ~\figref{fig:extra3}, we show more results with Ours using~\modelname. %

\section{Nearest neighbors of generated images in FFHQ dataset}\label{apx:NN}

\begin{figure*}[t]
\centering
\setlength{\tabcolsep}{1pt}
\setlength\arrayrulewidth{1.5pt}
\begin{tabular}{c|ccccc}%
Generated &\multicolumn{5}{c}{Nearest Neighbors in FFHQ} \\

\begin{subfigure}{0.14\textwidth}\centering\includegraphics[width = \textwidth]{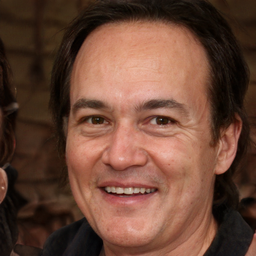}\end{subfigure}&
\begin{subfigure}{0.14\textwidth}\centering\includegraphics[width = \textwidth]{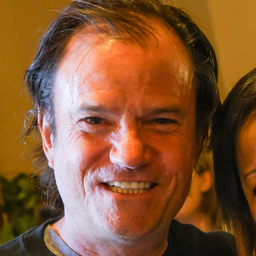}\end{subfigure}&
\begin{subfigure}{0.14\textwidth}\centering\includegraphics[width = \textwidth]{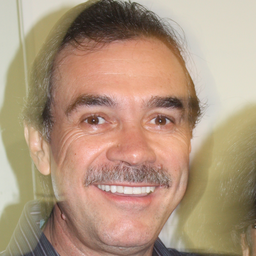}\end{subfigure}&
\begin{subfigure}{0.14\textwidth}\centering\includegraphics[width = \textwidth]{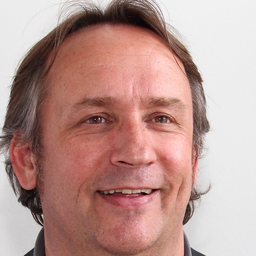}\end{subfigure}&
\begin{subfigure}{0.14\textwidth}\centering\includegraphics[width = \textwidth]{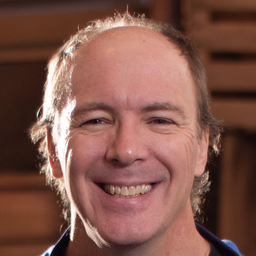}\end{subfigure}&
\begin{subfigure}{0.14\textwidth}\centering\includegraphics[width = \textwidth]{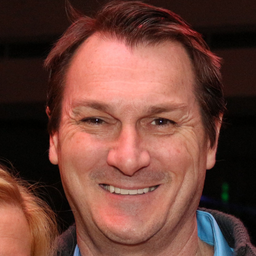}\end{subfigure}\\

\begin{subfigure}{0.14\textwidth}\centering\includegraphics[width = \textwidth]{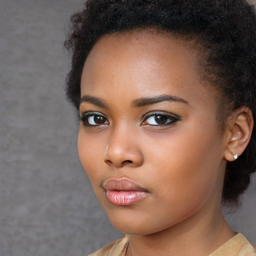}\end{subfigure}&
\begin{subfigure}{0.14\textwidth}\centering\includegraphics[width = \textwidth]{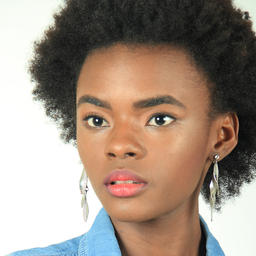}\end{subfigure}&
\begin{subfigure}{0.14\textwidth}\centering\includegraphics[width = \textwidth]{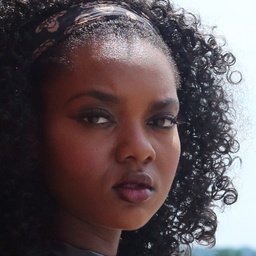}\end{subfigure}&
\begin{subfigure}{0.14\textwidth}\centering\includegraphics[width = \textwidth]{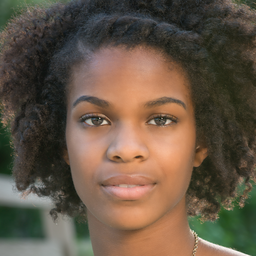}\end{subfigure}&
\begin{subfigure}{0.14\textwidth}\centering\includegraphics[width = \textwidth]{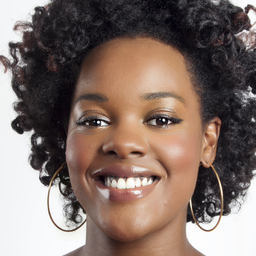}\end{subfigure}&
\begin{subfigure}{0.14\textwidth}\centering\includegraphics[width = \textwidth]{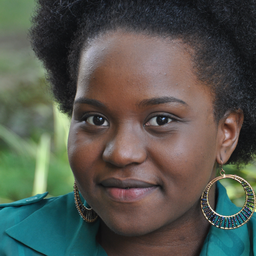}\end{subfigure}\\

\begin{subfigure}{0.14\textwidth}\centering\includegraphics[width = \textwidth]{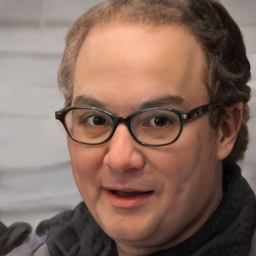}\end{subfigure}&
\begin{subfigure}{0.14\textwidth}\centering\includegraphics[width = \textwidth]{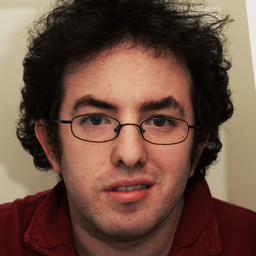}\end{subfigure}&
\begin{subfigure}{0.14\textwidth}\centering\includegraphics[width = \textwidth]{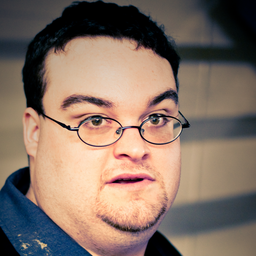}\end{subfigure}&
\begin{subfigure}{0.14\textwidth}\centering\includegraphics[width = \textwidth]{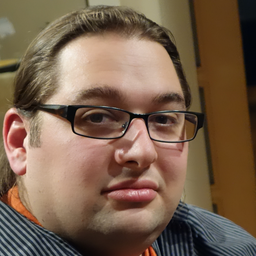}\end{subfigure}&
\begin{subfigure}{0.14\textwidth}\centering\includegraphics[width = \textwidth]{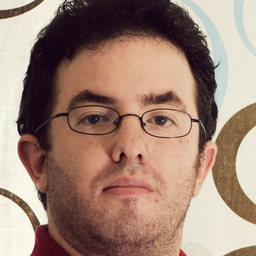}\end{subfigure}&
\begin{subfigure}{0.14\textwidth}\centering\includegraphics[width = \textwidth]{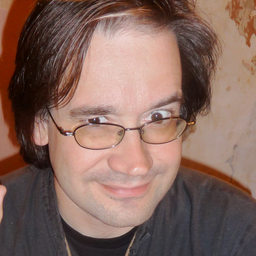}\end{subfigure}\\

\begin{subfigure}{0.14\textwidth}\centering\includegraphics[width = \textwidth]{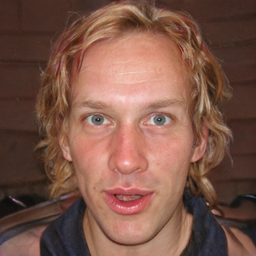}\end{subfigure}&
\begin{subfigure}{0.14\textwidth}\centering\includegraphics[width = \textwidth]{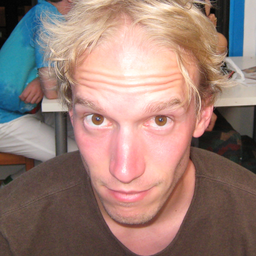}\end{subfigure}&
\begin{subfigure}{0.14\textwidth}\centering\includegraphics[width = \textwidth]{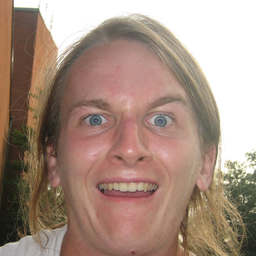}\end{subfigure}&
\begin{subfigure}{0.14\textwidth}\centering\includegraphics[width = \textwidth]{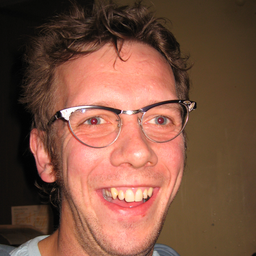}\end{subfigure}&
\begin{subfigure}{0.14\textwidth}\centering\includegraphics[width = \textwidth]{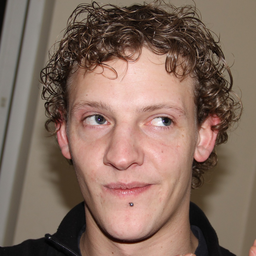}\end{subfigure}&
\begin{subfigure}{0.14\textwidth}\centering\includegraphics[width = \textwidth]{figures/supp/extra results/A person with a big forehead/nn/1_5.png}\end{subfigure}\\

\begin{subfigure}{0.14\textwidth}\centering\includegraphics[width = \textwidth]{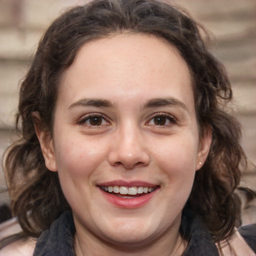}\end{subfigure}&
\begin{subfigure}{0.14\textwidth}\centering\includegraphics[width = \textwidth]{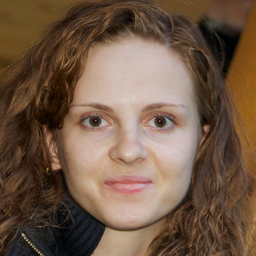}\end{subfigure}&
\begin{subfigure}{0.14\textwidth}\centering\includegraphics[width = \textwidth]{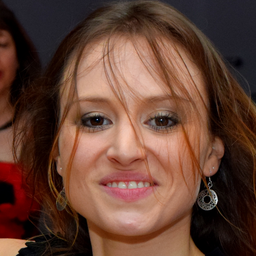}\end{subfigure}&
\begin{subfigure}{0.14\textwidth}\centering\includegraphics[width = \textwidth]{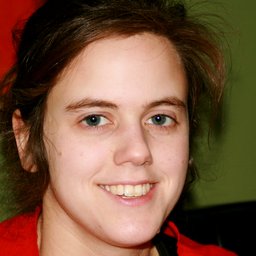}\end{subfigure}&
\begin{subfigure}{0.14\textwidth}\centering\includegraphics[width = \textwidth]{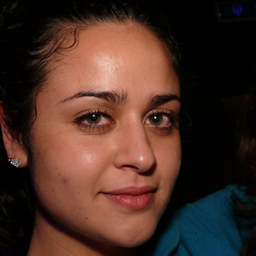}\end{subfigure}&
\begin{subfigure}{0.14\textwidth}\centering\includegraphics[width = \textwidth]{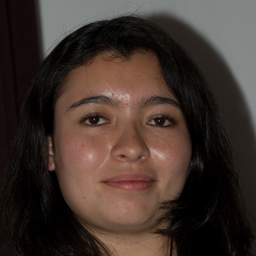}\end{subfigure}\\

\begin{subfigure}{0.14\textwidth}\centering\includegraphics[width = \textwidth]{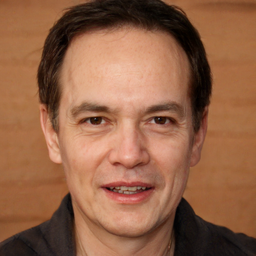}\end{subfigure}&
\begin{subfigure}{0.14\textwidth}\centering\includegraphics[width = \textwidth]{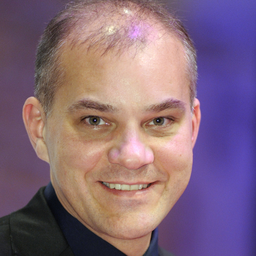}\end{subfigure}&
\begin{subfigure}{0.14\textwidth}\centering\includegraphics[width = \textwidth]{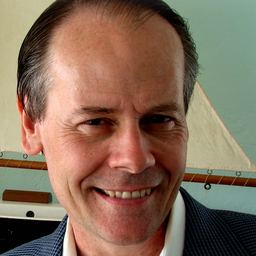}\end{subfigure}&
\begin{subfigure}{0.14\textwidth}\centering\includegraphics[width = \textwidth]{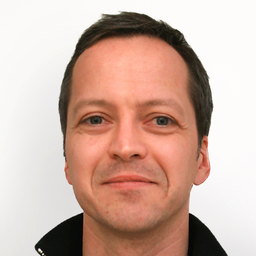}\end{subfigure}&
\begin{subfigure}{0.14\textwidth}\centering\includegraphics[width = \textwidth]{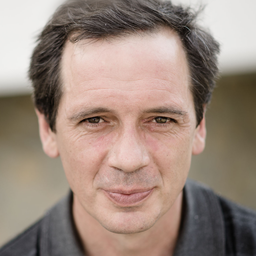}\end{subfigure}&
\begin{subfigure}{0.14\textwidth}\centering\includegraphics[width = \textwidth]{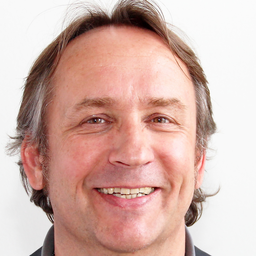}\end{subfigure}\\

\begin{subfigure}{0.14\textwidth}\centering\includegraphics[width = \textwidth]{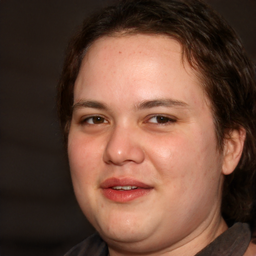}\end{subfigure}&
\begin{subfigure}{0.14\textwidth}\centering\includegraphics[width = \textwidth]{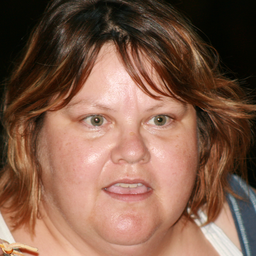}\end{subfigure}&
\begin{subfigure}{0.14\textwidth}\centering\includegraphics[width = \textwidth]{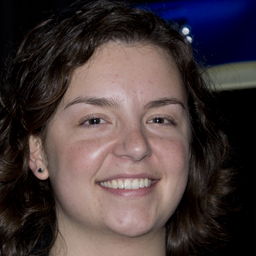}\end{subfigure}&
\begin{subfigure}{0.14\textwidth}\centering\includegraphics[width = \textwidth]{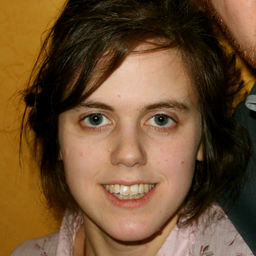}\end{subfigure}&
\begin{subfigure}{0.14\textwidth}\centering\includegraphics[width = \textwidth]{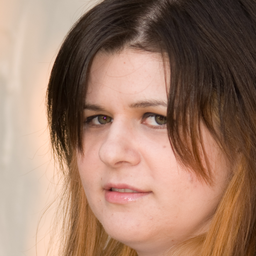}\end{subfigure}&
\begin{subfigure}{0.14\textwidth}\centering\includegraphics[width = \textwidth]{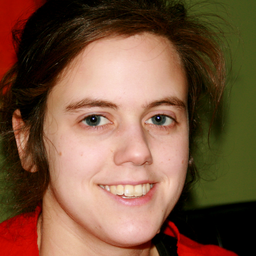}\end{subfigure}\\

\begin{subfigure}{0.14\textwidth}\centering\includegraphics[width = \textwidth]{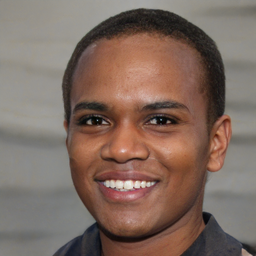}\end{subfigure}&
\begin{subfigure}{0.14\textwidth}\centering\includegraphics[width = \textwidth]{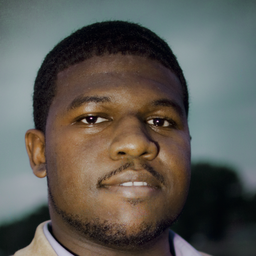}\end{subfigure}&
\begin{subfigure}{0.14\textwidth}\centering\includegraphics[width = \textwidth]{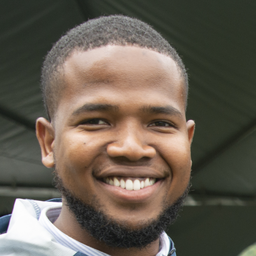}\end{subfigure}&
\begin{subfigure}{0.14\textwidth}\centering\includegraphics[width = \textwidth]{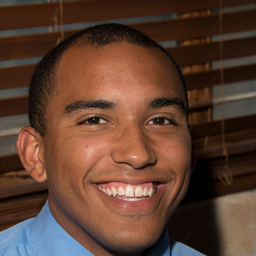}\end{subfigure}&
\begin{subfigure}{0.14\textwidth}\centering\includegraphics[width = \textwidth]{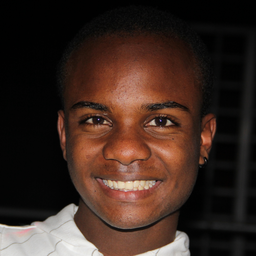}\end{subfigure}&
\begin{subfigure}{0.14\textwidth}\centering\includegraphics[width = \textwidth]{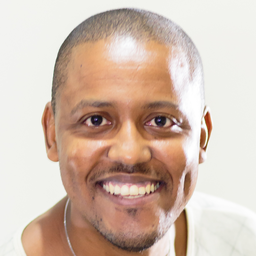}\end{subfigure}

\end{tabular}
\vspace{0.02cm}
\caption{Nearest neighbors of generated images in FFHQ dataset, drawn by computing CLIP similarity, on the text prompt ``A person with a big forehead". Our model creates novel faces with novel identities each time even given the same text prompt.}
\vspace{-0.3cm}
\label{fig:neighbors}
\end{figure*}

\begin{figure*}[t]
\centering
\setlength{\tabcolsep}{0pt}
\renewcommand{\arraystretch}{0.0}
\begin{tabular}{cccc}

\multicolumn{4}{c}{\small 
    \begin{tabular}{c}
    A young woman with colorful hair.
  \end{tabular}} \\
\includegraphics[width = .21\textwidth]{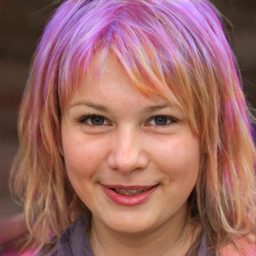}&
\includegraphics[width = .21\textwidth]{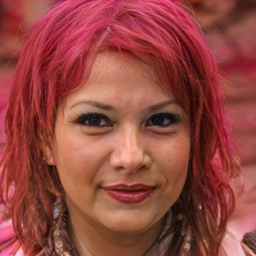}&
\includegraphics[width = .21\textwidth]{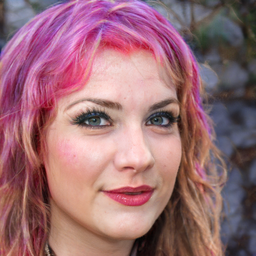}&
\includegraphics[width = .21\textwidth]{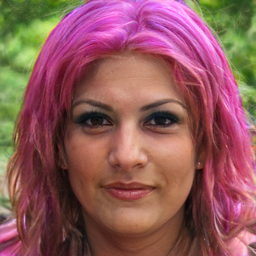}\\
\includegraphics[width = .21\textwidth]{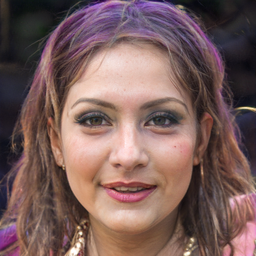}&
\includegraphics[width = .21\textwidth]{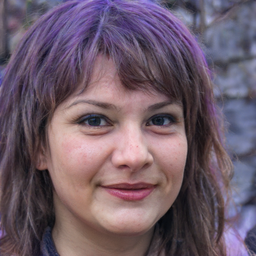}&
\includegraphics[width = .21\textwidth]{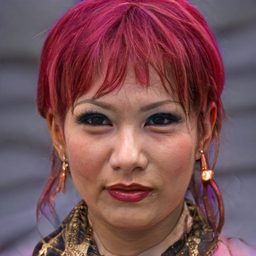}&
\includegraphics[width = .21\textwidth]{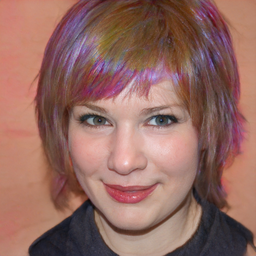}\\ [5pt]

\multicolumn{4}{c}{\small 
    \begin{tabular}{c}
    An Asian with long hair.
  \end{tabular}} \\
\includegraphics[width = .21\textwidth]{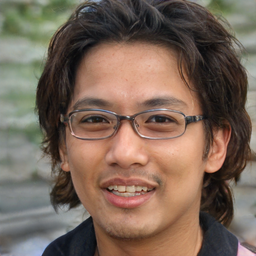}&
\includegraphics[width = .21\textwidth]{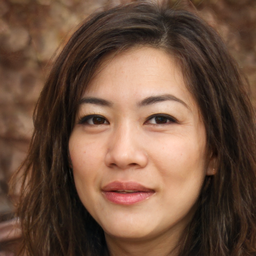}&
\includegraphics[width = .21\textwidth]{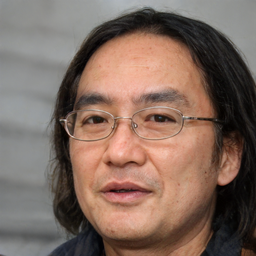}&
\includegraphics[width = .21\textwidth]{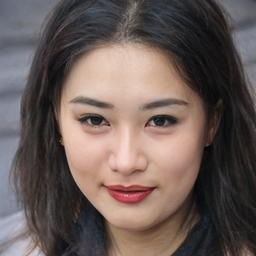}\\
\includegraphics[width = .21\textwidth]{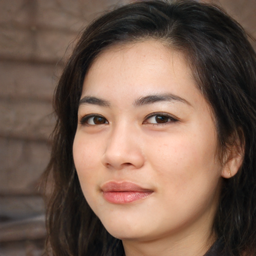}&
\includegraphics[width = .21\textwidth]{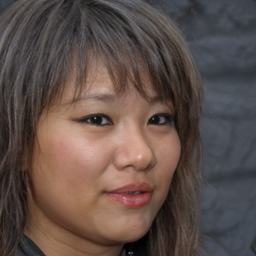}&
\includegraphics[width = .21\textwidth]{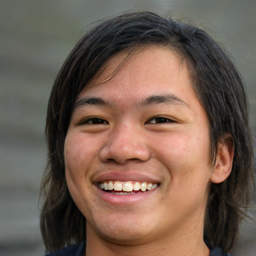}&
\includegraphics[width = .21\textwidth]{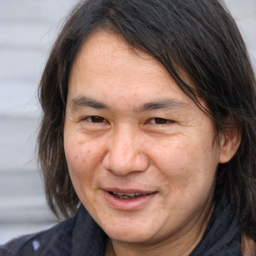}\\ [5pt]

\multicolumn{4}{c}{\small 
    \begin{tabular}{c}
    A person with a big forehead.
  \end{tabular}} \\
\includegraphics[width = .21\textwidth]{figures/supp/extra results/A person with a big forehead/1.png}&
\includegraphics[width = .21\textwidth]{figures/supp/extra results/A person with a big forehead/4.png}&
\includegraphics[width = .21\textwidth]{figures/supp/extra results/A person with a big forehead/3.png}&
\includegraphics[width = .21\textwidth]{figures/supp/extra results/A person with a big forehead/2.png}\\
\includegraphics[width = .21\textwidth]{figures/supp/extra results/A person with a big forehead/10.png}&
\includegraphics[width = .21\textwidth]{figures/supp/extra results/A person with a big forehead/6.png}&
\includegraphics[width = .21\textwidth]{figures/supp/extra results/A person with a big forehead/7.png}&
\includegraphics[width = .21\textwidth]{figures/supp/extra results/A person with a big forehead/12.png}

\end{tabular}
\caption{Additional results for Ours. The text prompt on top is used to generate the eight images below it.
}
\label{fig:extra3}
\end{figure*}

In~\figref{fig:neighbors} We show the nearest FFHQ neighbors of images generated by Ours. The left column contains the outputs from the last caption in~\figref{fig:extra3}. We use CLIP to find their top 5 nearest neighbors from FFHQ dataset in CLIP feature space. The neighbors are on the same row and to the right of the output image. The neighbors are ranked from left to right with top 1 nearest neighbors on the left. As can be seen, the outputs are not duplicates of images from FFHQ but novel faces with novel identities. 

\section{Text prompts of~\figref{fig:pose_compare} in the main body}\label{apx:prompts}
From left to right:
\begin{enumerate}
    \item ``A bald Asian man with beard."
    \item ``A photo of a smiling Asian young man with curly hair."
    \item ``A blond woman with wavy long hair wearing makeup."
\end{enumerate}



\clearpage

\end{document}